\documentclass[final,3p,times]{elsarticle}
\usepackage{amsmath,amsfonts}
\usepackage{algorithmic}
\usepackage{algorithm}
\usepackage{array}
\usepackage[caption=false,font=normalsize,labelfont=sf,textfont=sf]{subfig}
\usepackage{textcomp}
\usepackage{stfloats}
\usepackage{url}
\usepackage{verbatim}
\usepackage{graphicx}
\usepackage{wrapfig} 
\usepackage{booktabs} 
\usepackage{bookmark}
\usepackage{pifont}
\usepackage{bbm}
\usepackage{soul}
\usepackage{color, xcolor}
\usepackage{placeins}
\usepackage{float}
\usepackage{stfloats}
\usepackage{lipsum}
\usepackage{xtab}
\usepackage{bm} 
\usepackage{multicol} 
\usepackage{multirow}

\journal{Robotics and Computer-Integrated Manufacturing}

\newcommand{\fn}[1]{\textcolor{blue}{Fig.~\ref{#1}}}
\newcommand{\tn}[1]{Table~\ref{#1}} 

\newcommand{\sn}[1]{Section~\ref{#1}} 

\newcommand{\en}[1]{\textcolor{blue}{Eq.~(\ref{#1})}} 
 
\newcommand{\blue}[1]{\textcolor{blue}{#1}}

\soulregister{\cite}7 
\soulregister{\citep}7 
\soulregister{\citet}7 
\soulregister{\ref}7 
\soulregister{\pageref}7 

\usepackage{balance}

\makeatletter
\newenvironment{breakablealgorithm}
  {
   \begin{center}
     \refstepcounter{algorithm}
     \hrule height.8pt depth0pt \kern2pt
     \renewcommand{\caption}[2][\relax]{
       {\raggedright\textbf{\ALG@name~\thealgorithm} ##2\par}%
       \ifx\relax##1\relax 
         \addcontentsline{loa}{algorithm}{\protect\numberline{\thealgorithm}##2}%
       \else 
         \addcontentsline{loa}{algorithm}{\protect\numberline{\thealgorithm}##1}%
       \fi
       \kern2pt\hrule\kern2pt
     }
  }{
     \kern2pt\hrule\relax
   \end{center}
  }
\makeatother
\begin{document}

\begin{frontmatter}
\title{Knowledge-Guided Hierarchical Policy Learning for High-Precision Cylindrical Assembly under Tight Tolerances}

\author{Binbin Lian, Xinyu Liu}

\author{Tao Sun\corref{cor1}} 
\ead{stao@tju.edu.cn}
\affiliation{organization={Key Laboratory of Mechanism Theory and Equipment Design of Ministry of Education},
            addressline={Tianjin University}, 
            city={Tianjin},
            postcode={300350}, 
            country={China}}

\cortext[cor1]{Corresponding author}


\begin{abstract}
  A hybrid hierarchical learning framework is proposed to achieve high-precision assembly of $\Phi$170mm cylindrical components with tolerance of 0.1mm. The lower-level network integrates expert experience through Behavior Cloning (BC), giving the robot human-like intuition, and incorporates the Twin Delayed Deep Deterministic Policy Gradient (TD3) algorithm to enhance training stability and robustness. The upper-level network dynamically adjusts the lower-level decisions based on heuristic rules, ensuring flexibility in operations. A simulated model is constructed to learn before transferring to real world. An efficient and safe training is allowed. Comparisons show that the reward curve converges within 500 episodes, indicating high learning efficiency. It also demonstrates better adaptability to initial conditions and pose errors, achieving satisfactory success rates even under extreme conditions. Moreover, the method exhibits good stability under Gaussian noise interference. In the real world, the assembly trajectory of the cylindrical segment shows smoother motion and less fluctuation.
\end{abstract}

\begin{keyword}
  Parallel robot; Peg-in-hole assembly; Imitation learning; Reinforcement learning
\end{keyword}

\end{frontmatter}

\section{Introduction}\label{Section: Introduction}

Assembly of large size and heavy components in areas like automobiles and aircraft still largely relies on manual labor \cite{add1}. Thanks to the flexibility and adaptivity, workers could conduct assembling task with high quality after skill training. But it is labor intensive. Especially for large and heavy components with tight tolerance, consistency and accuracy of manual assembling cannot be assured \cite{add2,add3}. Owing to high precision and efficient execution capability, robotic assembly has become a trend \cite{1,2,3}. But for assembly in unstructured environment with uncertain variations, complex models are demanded for the robot to improve robustness to environmental noise and disturbances \cite{4}. Traditional modeling and control methods sometimes cannot effectively address this problem due to the lack of pre-knowledge and potential uncertainties. To address the shortcomings in traditional robotic assembly systems, Zeng et al. introduced a method for large component assembly using iGPS and laser ranging \cite{5}. This method is efficient, but it has limited accuracy in specific directions. Li et al. developed a force-controlled mobile handling system for aircraft assembly \cite{6}. This system enhances flexibility and efficiency, but it lacks adaptability to complex assembly features and high-precision tasks. Zheng et al. proposed a method for measuring the pose of large cylinders using binocular vision and prior data \cite{7}. It achieves precise positioning but is insufficiently adaptable to dynamic and complex environments.

In recent years, machine learning, particularly reinforcement learning (RL), has significantly progressed in robotic control. RL, through interaction with environment, can autonomously learn optimal strategies without explicit models, demonstrating strong adaptability and efficient exploration capabilities \cite{8,9,10}. For example, Wu et al. proposed a prioritized dueling deep Q network (DQN). They combined it with long short-term memory (LSTM) networks to achieve millimeter-level precision in automated assembly tasks \cite{11}. Hwangbo et al. introduced an off-policy, model-free deep RL method applicable to unstructured environments, enabling robot to perform compliant assembly tasks without manual tuning \cite{12}. Kozlovsky et al. employed model-free RL to learn asymmetric impedance policies for small-scale peg-in-hole assembly, reducing the need to learn full reference trajectories \cite{add4}. Although RL exhibits great potential, most existing studies focus on small-scale peg-in-hole assembly with robotic manipulators. However, large-size components with tight tolerances pose significant challenges for RL-based assembly, and related research remains scarce. In particular, the target components in this work typically exceed $\Phi$150mm in diameter with an assembly tolerance of only 0.1mm. In the meantime, there are multiple connecting holes on each part requiring to be aligned. {Lutz et al. proposed a model-free RL approach for adaptive shape control in aircraft fuselage assembly, building a reinforcement learning environment via finite element simulation and modifying the PPO algorithm to optimize actuator layout and force for shape adjustment \cite{add5}. This method enables high-precision fuselage shape control and excellent assembly gap reduction, yet it relies on complex finite element simulation modeling with high implementation complexity and time costs. Liang et al. proposed a visual-haptic fusion framework for large-diameter peg-in-hole assembly, integrating an improved U-Net–FPN for peg localization and DDPG-based force-aware control for compliant insertion \cite{add6}. This approach achieves stable assembly with good deviation tolerance, but its visual recognition relies on a small dataset with limited environmental variability, and fixed camera configuration plus low control efficiency restrict dynamic large-component assembly applications. Liu et al. proposed a DRL-based assembly path planning method for the lateral thrust device of pulse solid rocket motors \cite{add7}. This method leverages finite element simulation to obtain O-ring damage data and adopts dueling DDQN to generate high-precision assembly paths, effectively reducing O-ring damage, yet it is only validated in a single plane with limited degrees of freedom. For the assembly of large-size components with tight tolerances and multi-hole alignment requirements, RL algorithms face significant inefficiency in exploring assembly features and determining appropriate actions, and the probability of assembly failure remains high. In addition, RL requires large amount of training data. Each action and decision indicate one round of robot execution. Assembling learning is inefficient because of frequent robot manipulation \cite{add8,add9}. Moreover, improper decision may lead to collision or unnecessary damage to the component in real world due to large inertia. 

Since applying RL alone has slow convergence and low exploration efficiency \cite{13,14,15}, combining RL and other learning algorithms has become a promising solution. For instance, imitation learning allows robot to quickly learn useful information from human demonstrations and convert it into knowledge models, thereby obtaining high-level control strategies and accelerating RL process. Triantafyllidis et al. proposed a Robot Operation Network (ROMAN). Behavior cloning, imitation learning, and RL are integrated to enable robotic systems to perform a variety of sequential tasks with operational skills and strong fault recovery capabilities \cite{16}. Cho et al. proposed a framework based on imitation learning and teaching to make robot learn, improve, and generalize motor skills. The robot learns initial motor skills from human demonstrations. It improves the skills through RL and further applies to motors with other shapes \cite{17}. Li et al. proposed a goal-conditioned self-imitation RL method (GCDI) for 3C industry flexible flat cable assembly, integrating hindsight experience replay and self-imitation learning to tackle sparse rewards and limited learning conditions \cite{add10}. To further address complex execution procedures in RL-based skill learning, a hierarchical RL approach where high-level networks output action bounds to guide low-level actions, reducing low-level network learning difficulty and enhancing real-world adaptability and control accuracy. Macpherson et al. applied high-level policy to adjust low-level amplitude parameters for prediction and control of complex behaviors \cite{18}. Wang et al. proposed a two-layer RL learning scheme to generate gait patterns for modular robots. A high-level network is utilized to modulate low-level actions, achieving smooth motion in gait control efficiently \cite{19}. Huang et al. proposed a three-layer hierarchical RL architecture for hexapod robots, where the high-level PPO controller optimizes mid-level CPG parameters and low-level mapping function parameters, enabling the robot to adapt stably to complex terrains \cite{add11}. As for efficiency and safety issues in early-stage RL exploration, implementing learning process firstly in virtual environment before transferring to physical system ensures safety and improves efficiency. Ju et al. proposed a deep RL method, where control strategies are trained in a simulated environment and then transferred to real world improving robot performance \cite{20}. Peng et al. trained the robot control strategy by domain randomization in simulation. Control strategies are more robust in real-world tasks \cite{21}. Yan et al. constructed a multi-category peg-in-hole assembly virtual environment set with mathematical and physical force contact models, training the adaptive meta policy learning algorithm in it to obtain a generalized assembly skill model that performs excellently in real-world single, dual, and triple peg-in-hole assembly without additional training \cite{add12}.

Based on the above ideas, a hybrid hierarchical learning (HL) framework is proposed for docking and assembling cylindrical components over $\Phi$150mm with tolerance 0.1 mm. The HL framework combines LSTM, Behavior Cloning (BC) and Twin Delayed Deep Deterministic Policy Gradient (TD3). Robot actions follow assembling procedure that first adjust posture and then translate to align holes. In addition, a simulated environment is developed for robot learning instead of implementing it directly in real world. Contributions are summarized as follows.
\begin{enumerate}
  \item{Lower-level network of the learning framework is composed of BC and TD3. BC embeds knowledge-guided heuristics from expert demos to give the robot human-like intuition. An initial policy is then offered to TD3, which learns docking skills more stable and efficiently. Constraints are set in the penalty terms of reward function to avoid potential risks and improve learning efficiency.}
  \item{Upper-level network is designed to select the lower-level network and control the step size of robot execution for fast approaching and fine adjustment. The LSTM dynamically guides the lower-level network based on heuristic rules obtained from an expert dataset.}
  \item{A near-real simulation model is constructed. Relative postures between two parts are obtained by initial setting and robot kinematics. The learning outcome is directly transferred to the real world where system state is captured by binocular camera. Efficient and safe learning is allowed in the simulated environment.}
\end{enumerate}
The remainder of this paper is organized as follows: \sn{Section: PRELIMINARY} reviews related works on cylindrical segment assembly. \sn{Section: METHOD} presents the hybrid hierarchical learning framework and algorithm. \sn{Section: SIMULATION} describes the simulation environment and evaluates the proposed framework. \sn{Section: EXPERIMENTS AND ANALYSIS} discusses the transfer method from simulation to reality and presents experimental results in real world. \sn{Section: DISCUSSION} provides a discussion on the findings and their implications. Finally, \sn{Section: CONCLUSION} concludes the study and discusses future work.

\section{Preliminary}\label{Section: PRELIMINARY}
\subsection{Cylinder segment assembly}
The setup for cylinder segment assembly is shown in \fn{Fig_1}. Segment I is installed on a six-degree-of-freedom parallel robot while Segment II is mounted on a fixture. Assembling features include peg-in-hole and connecting holes. Diameter of shaft on Segment I is $\Phi$169.90 mm and that of the hole on Segment II is $\Phi$170.00 mm. The eight $\Phi$6.00 mm connecting holes are along the radius directions of shaft in Segment I and symmetrically distributed. Similarly, eight $\Phi$6.00 mm connecting holes are on the hole in Segment II. The parallel robot carries Segment I to assemble with Segment II. Shaft and hole are expected to be assembled and the connecting holes are aligned. Perception is realized by binocular camera. The gap between cylinder segments is only 0.1 mm, making the system very sensitive to errors and uncertainties.

\begin{figure}[h]
    \centering
    \includegraphics[width=0.5\columnwidth]{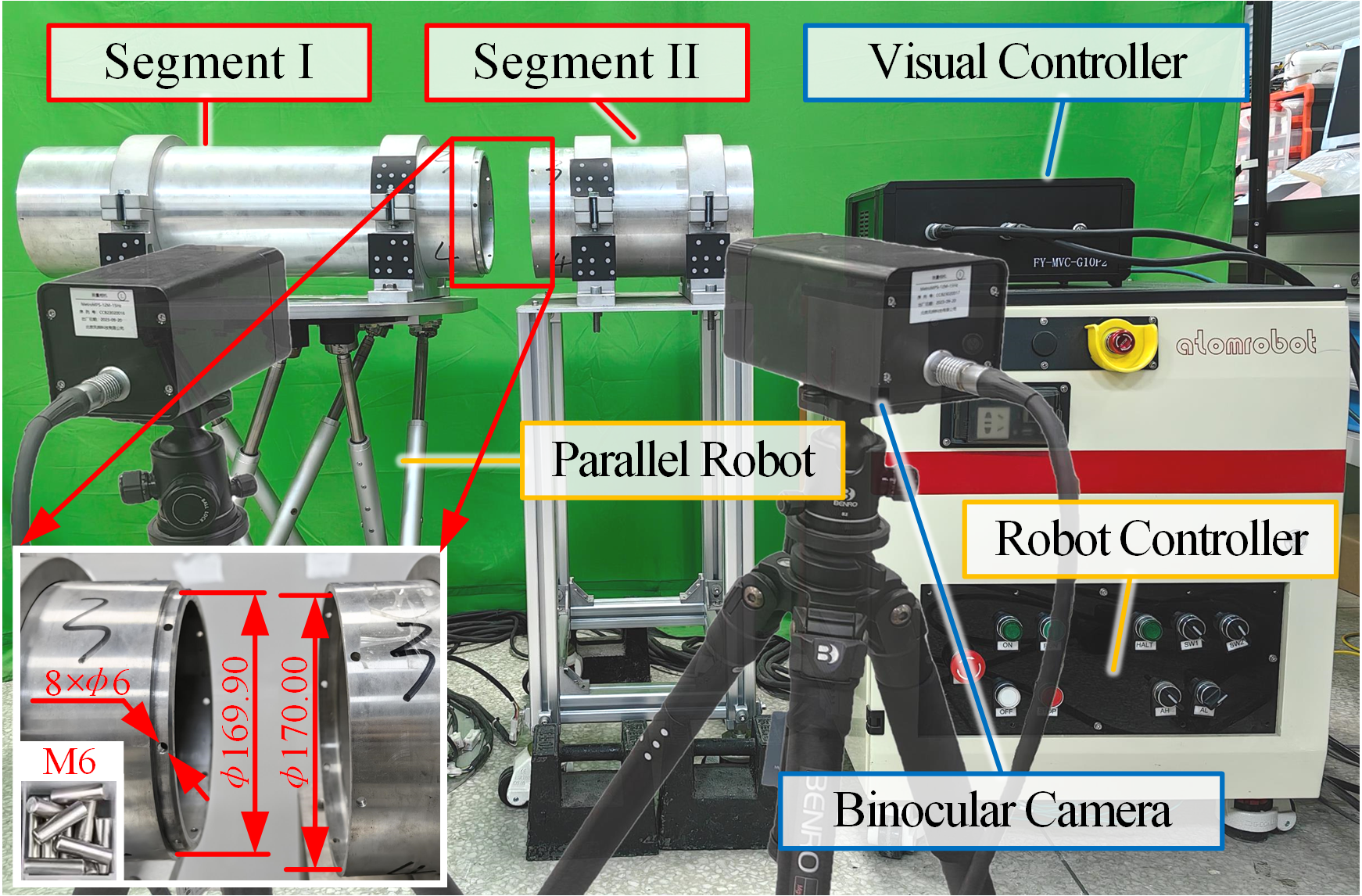}
    \caption{Cylindrical segment assembly system and docking features.}
    \label{Fig_1}
\end{figure}

\subsection{Process parameter optimization}
\subsubsection{Assembly system coordinate system definition}
This study establishes a coordinate system for the assembly system by employing a binocular camera system and encoded fiducial points, effectively parameterizing the tight tolerance assembly problem. The binocular camera system uses the MetroMPS-D12 vision measurement camera, with a measurement range of 2-10 m and a measurement accuracy of $12\ \mu\text{m} + 12\ \mu\text{m/m}$. The camera system is calibrated with a calibration board prior to each use. This approach not only facilitates the subsequent modeling and analysis but also provides a structured framework for the spatial relationships within the assembly process. The coordinate systems are illustrated in \fn{Fig_2}.

Frame $\{W\}$ serves as the base coordinate system, established based on the left camera of the binocular camera system. All other frames are measured relative to frame $\{W\}$. Frame $\{C_1\}$ and frame $\{C_2\}$ located on the fixtures of Segment I and II. They are defined by three non-collinear encoded points captured directly by the binocular camera. These frames are used to determine the positions of other frames during assembly. Frame $\{M\}$ and frame $\{F\}$ are the coordinate systems of Segment I and II. Frame $\{P\}$ is the coordinate system on the moving platform of the parallel robot. During the assembly, the spatial relations between frame $\{C_1\}$ and frame $\{M\}$, frame $\{C_1\}$ and frame $\{P\}$, as well as frame $\{C_2\}$ and frame $\{F\}$, remain constant. The following transformation matrices are applied:

\begin{equation}
\label{Eq1}
\left\{\begin{matrix}M={\mathbf{T}}_{{C}_{1}\_ M}\cdot{C}_{1}
\\ F={\mathbf{T}}_{{C}_{2}\_ F}\cdot{C}_{2}
\\ P={\mathbf{T}}_{{C}_{1}\_ P}\cdot{C}_{1}
\end{matrix}\right.
\end{equation}

\noindent where $\mathbf{T}_{C_1\_ M}$, $\mathbf{T}_{C_1\_ P}$ and $\mathbf{T}_{C_2\_ P}$ denote the transformation matrices measured by camera.

Frame $\{O\}$ is on the fixed base of parallel robot. It is derived through frame $\{P\}$ using forward kinematics of parallel robot. The required joint actuation values for the parallel robot are calculated using inverse kinematics. By recording position and orientation of frames during the human-robot collaborative assembling, expert data is collected.

\begin{figure}[h]
    \centering
    \includegraphics[width=0.5\columnwidth]{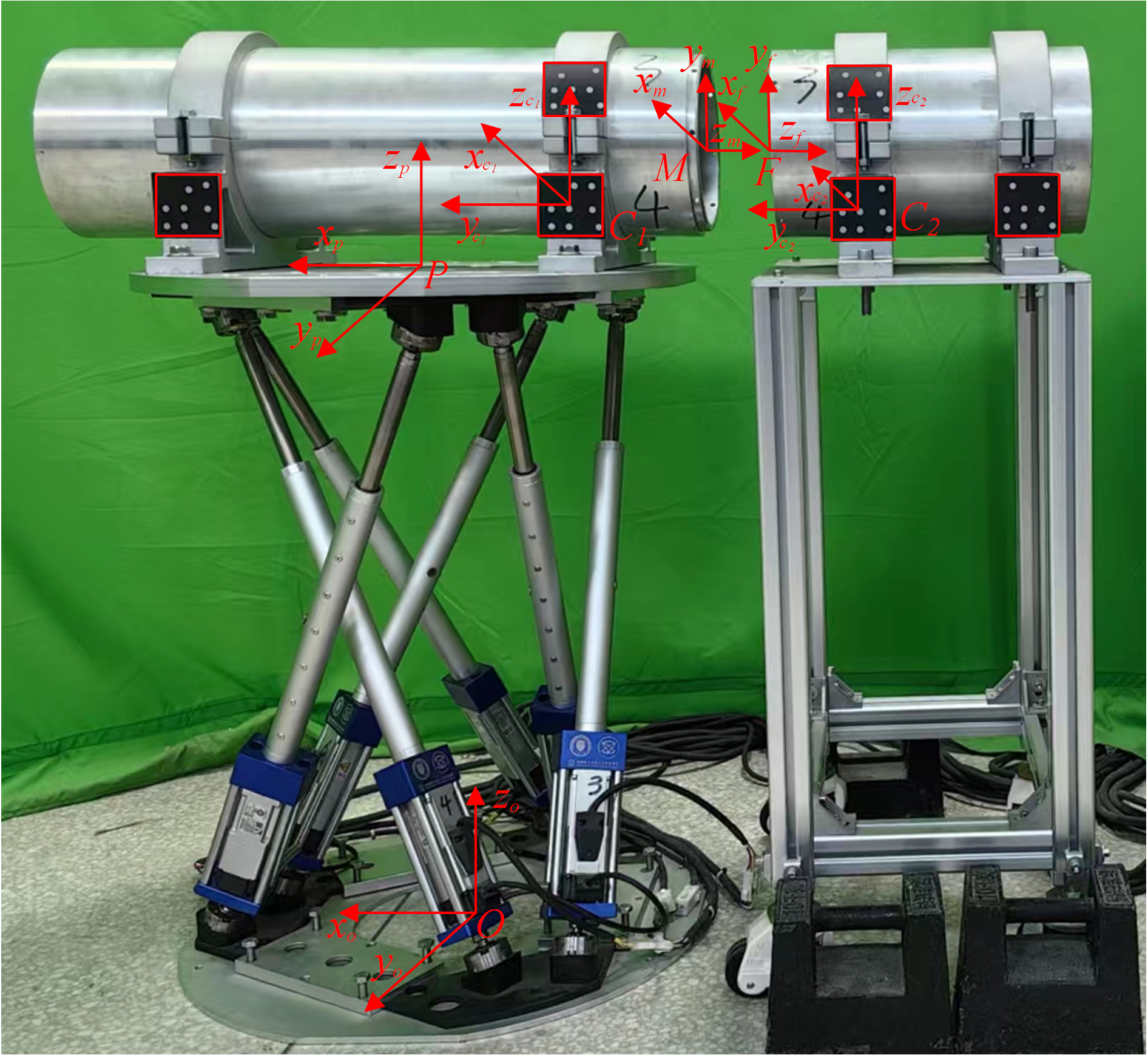}
    \caption{Assembly system coordinate system.}
    \label{Fig_2}
\end{figure}

\subsubsection{Parallel robot system}
We deploy a 6-DoF parallel robot for the precision assembly task. The 6-DoF parallel robot is as shown in \fn{Fig_3}, whose topological layout is 6-SPS. Herein, S and P are spherical joint and actuated prismatic joint. Point $A_i,B_i(i=1,2,...,6)$  are centers of S joints.

\begin{figure}[h]
    \centering
    \includegraphics[width=0.3\columnwidth]{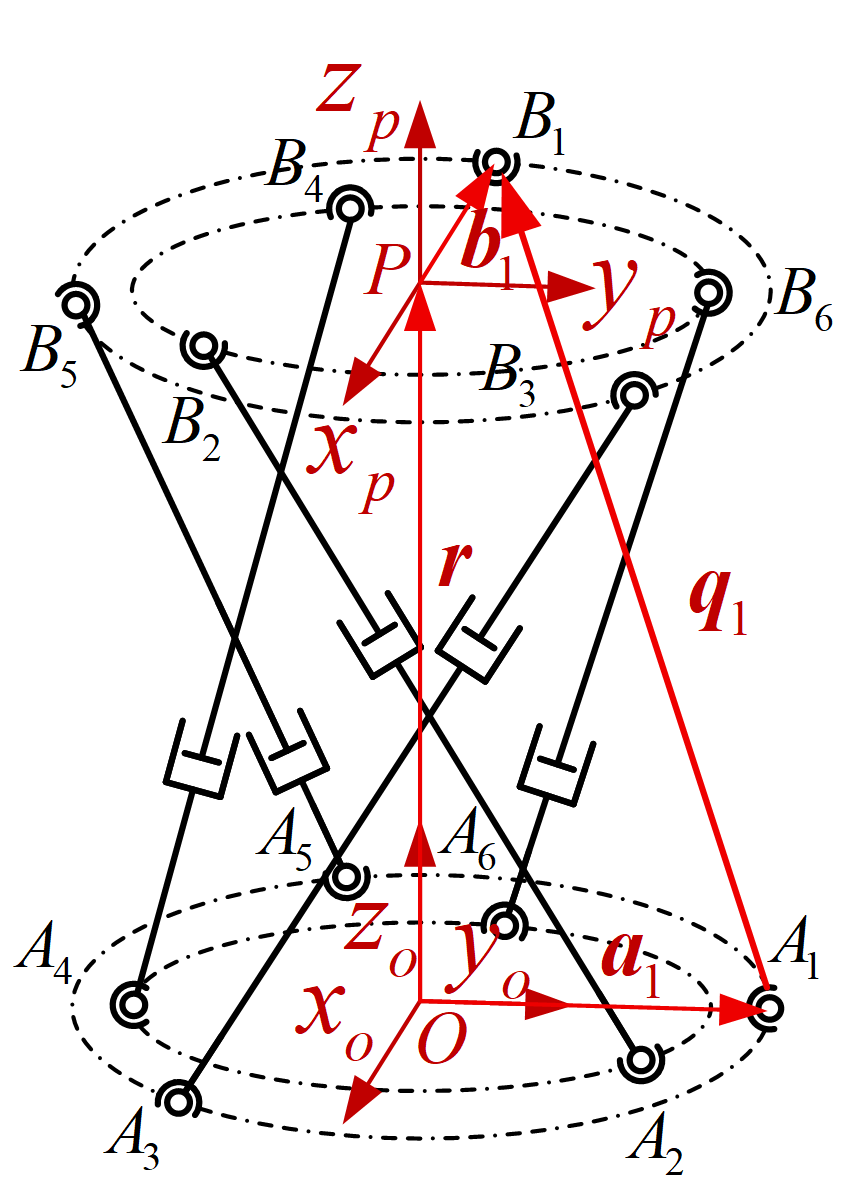}
    \caption{Robot mechanism diagram.}
    \label{Fig_3}
\end{figure}

In the context of any given kinematic chain, the construction of a closed-loop vector quadrilateral is feasible, yielding the subsequent relationship:

\begin{equation}
\label{Eq2}
\textbf{\textit{q}}_i=\textbf{\textit{r}}+\mathbf{R}\textbf{\textit{b}}_i-\textbf{\textit{a}}_i(i=1,2,...,6)
\end{equation}

\noindent where $\textbf{\textit{r}}$ is the vector directed from point $O$ to point $P$. $\textbf{\textit{a}}_i$ is the vector from point $O$ to point $A_i$ within frame $\{O\}$. $\textbf{\textit{b}}_i$ is the vector from point $P$ to point $B_i$ within frame $\{P\}$. $\textbf{\textit{q}}_i$ is the positional vector of the moving link. $\mathbf{R}$ is the rotational transformation that aligns frame $\{P\}$ with frame $\{O\}$.

\subsubsection{Kinematic model}
The integration of a robotic system into a hybrid hierarchical learning framework necessitates an accurate kinematic model. It translates the robot's physical capabilities into a format that the learning framework can use to control movements effectively. This integration ensures precise robot operations. For inverse kinematics, the method outlined in \cite{22} is employed. In this section, we present an exhaustive exposition on forward kinematics.

The forward kinematics of the parallel robot is a set of highly nonlinear equations with six unknown parameters, solved using the Newton-Raphson numerical method, as described in \cite{23}. 

Transformation of \en{Eq2} yields the following:

\begin{equation}
\textbf{\textit{F}}_i(\textbf{\textit{X}})=-\textbf{\textit{q}}_i^2+\begin{bmatrix}\textbf{\textit{r}}+\mathbf{R}\textbf{\textit{b}}_i-\textbf{\textit{a}}_i
\end{bmatrix}{}^{\mathrm{T}}
\begin{bmatrix}
\textbf{\textit{r}}+\mathbf{R}\textbf{\textit{b}}_i-\textbf{\textit{a}}_i
\end{bmatrix}=0
\end{equation}

\noindent where $\textbf{\textit{X}}=[x\ y\ z\ \alpha\ \beta \ \gamma ]^\mathrm{T}$, is the end-effector pose of the platform.

\begin{equation}
\textbf{\textit{F}}(\textbf{\textit{X}})=[\textbf{\textit{F}}_1\ \textbf{\textit{F}}_2\ \textbf{\textit{F}}_3\ \textbf{\textit{F}}_4\ \textbf{\textit{F}}_5\ \textbf{\textit{F}}_6]^\mathrm{T}
\end{equation}

\noindent where $\textbf{\textit{F}}(\textbf{\textit{X}})$  is the structure obtained by integrating all the branches.

The Newton-Raphson method is employed to solve $\textbf{\textit{F}}(\textbf{\textit{X}})=0$, with the robot's initial pose serving as the initial solution $\textbf{\textit{X}}^{(0)}$ and the iteration tolerance denoted by $\varepsilon$. The solution is obtained by:

\begin{equation}
\textbf{\textit{F}}^{(0)}=\textbf{\textit{F}}(\textbf{\textit{X}}^{(0)})
\end{equation}

In accordance with the Newton iteration method, the solution is updated at the $k$-th iteration as:

\begin{equation}
\textbf{\textit{X}}^{(k+1)}=\textbf{\textit{X}}^{(k)}-(\mathbf{J}(\textbf{\textit{X}}^{(k)}))^{-1}\textbf{\textit{F}}(\textbf{\textit{X}}^{(k)})
\end{equation}

\noindent where $\mathbf{J}(\textbf{\textit{X}})=\frac{\partial \textbf{\textit{F}}}{\partial \textbf{\textit{X}}}=\begin{bmatrix}
 \frac{\partial \textbf{\textit{F}}_1}{\partial x}  \frac{\partial \textbf{\textit{F}}_1}{\partial y}  \frac{\partial \textbf{\textit{F}}_1}{\partial z}  \frac{\partial \textbf{\textit{F}}_1}{\partial \alpha }  \frac{\partial \textbf{\textit{F}}_1}{\partial \beta } \frac{\partial \textbf{\textit{F}}_1}{\partial \gamma }\\ 
....\end{bmatrix}$ is the Jacobian matrix.

The iteration continues until the condition $\begin{Vmatrix}\textbf{\textit{X}}^{(k+1)}-\textbf{\textit{X}}^{(k)}\end{Vmatrix}\leq \varepsilon $ is satisfied, at which point $\textbf{\textit{X}}^{(k+1)}$ is considered an approximate solution to $\textbf{\textit{F}}(\textbf{\textit{X}})=0$.

\subsubsection{Knowledge-guided expert acquisition}
Expert experience is acquired during human–robot collaborative assembly, as shown in \fn{Fig_4}. We employ the sensorless admittance-control framework proposed in \cite{24}. A parallel robot is modelled with finite and instantaneous screw theory. Motor currents are filtered by a Kalman estimator. Joint friction is identified offline with a LuGre model. External wrench applied by the operator is reconstructed without extra sensors. The operator grips Segment I and applies a guiding wrench to dock it into Segment II. A binocular camera captures the process at 4 Hz to provide expert data for BC learning.

\begin{figure}[h]
    \centering
    \includegraphics[width=0.45\columnwidth]{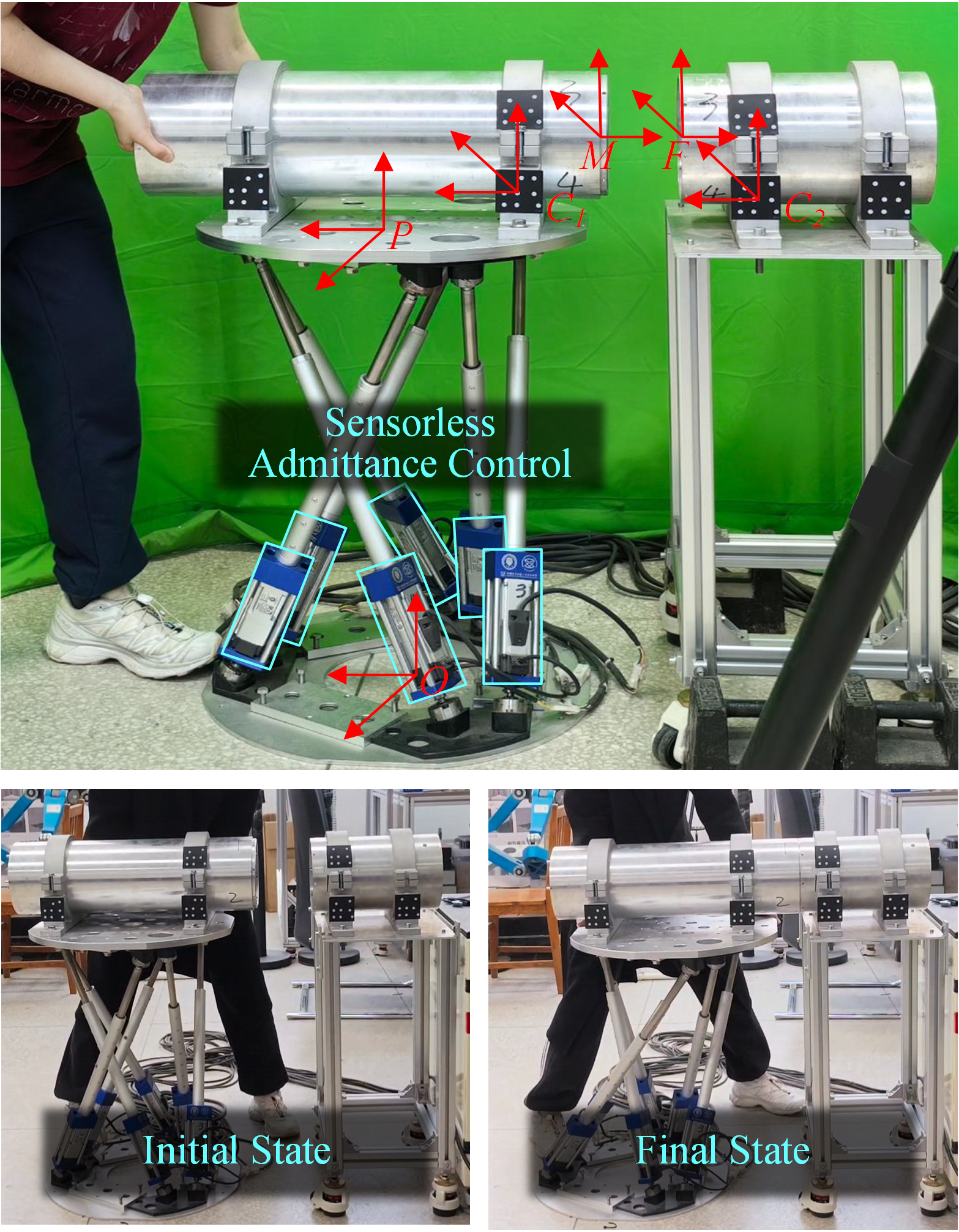}
    \caption{Acquisition of expert experience.}
    \label{Fig_4}
\end{figure}

\subsubsection{Collision detection mechanism}

To avoid component collisions and hardware damage risks during the assembly of cylindrical segments, a dedicated collision detection mechanism based on Oriented Bounding Box (OBB) is designed to judge whether subsequent states generated by actions will result in collisions, ensuring assembly safety and reliability.

For the geometric characteristics of cylindrical segments, minimal enclosing OBBs are constructed for Segment I and Segment II respectively. The central axis of the segment is taken as the $z$-axis of the OBB, and the $x,y$ axes are parallel to the radial direction of the segment's cross-section. OBB parameters include center coordinates $O_{I}$ and $O_{II}$, unit axis vectors ${x}_I,{y}_I,{z}_I$ for Segment I and ${x}_{II},{y}_{II},{z}_{II}$ for Segment II. The edge lengths along each axis are $W_I,H_I,D_I$ for Segment I, and $W_{II},H_{II},D_{II}$ for Segment II.

\begin{figure}[h]
	\centering
	\includegraphics[width=0.45\columnwidth]{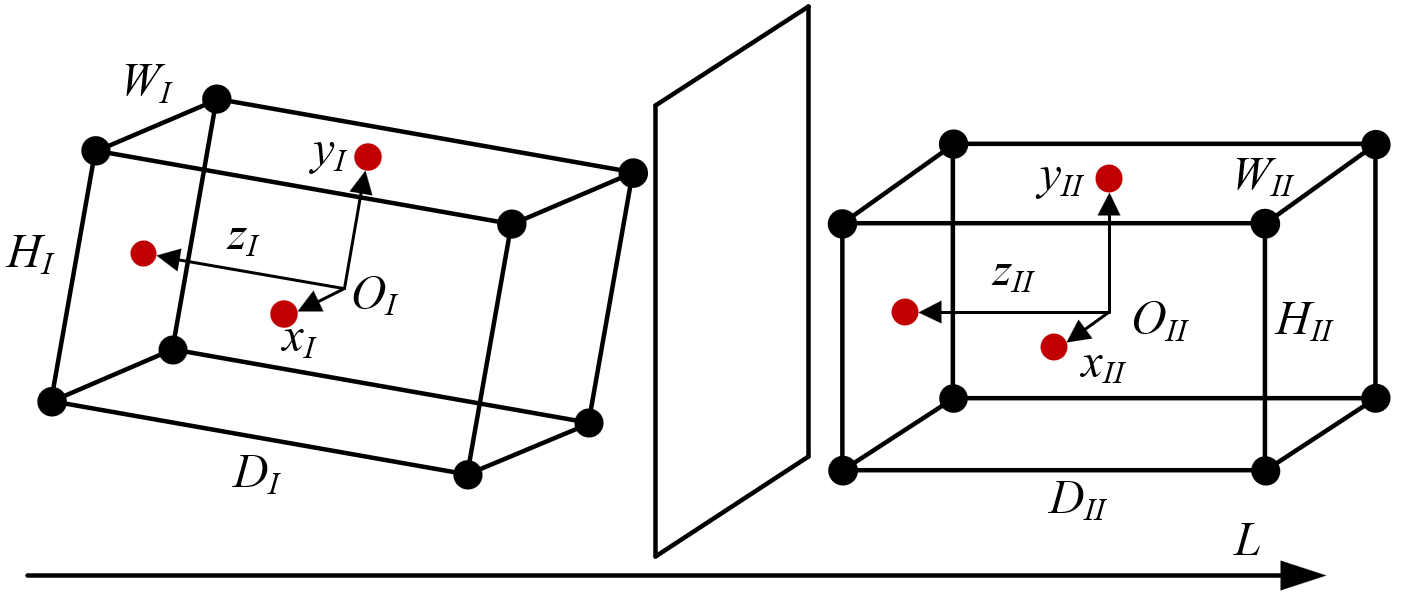}
	\caption{Bounding box collision detection diagram.}
	\label{Fig_add1}
\end{figure}

The core of collision detection adopts the separating plane criterion, as shown in \fn{Fig_add1}. A plane can completely separate the two OBBs, no collision is determined. Qtherwise, a collision risk exists. Its mathematical equivalent condition is as follows:
\begin{equation}
| L \cdot (O_{II} - O_I) | >  \frac{1}{2}(W_I|L \cdot x_I| + H_I|L \cdot y_I| + D_I|L \cdot z_I| + W_{II}|L \cdot x_{II}| + H_{II}|L \cdot y_{II}| + D_{II}|L \cdot z_{II}|)
\end{equation}

\noindent where $L \cdot (O_{II} - O_I)$ is the projection length of the line connecting the center points in the direction of the normal vector $L$.

There are fifteen candidate normal vectors, traversed by collision risk priority. First detect the axial directions aligned with the segment central axes $z_I,z_{II}$, two directions in total. Then detect the radial directions aligned with the radial axes $x_I,y_I,x_{II},y_{II}$, four directions in total. Finally detect the composite directions from the cross-products of the coordinate axes of the two segments, nine directions in total.

The detection process runs synchronously with the assembly motion. It samples the segment pose and calculates OBB parameters at each step. If the action execution state does not meet the required condition, a collision warning is issued. The current assembly action is canceled, the current state is read again, and the policy network generates a new action. If warnings occur more than 20 consecutive times, the current episode fails. The parallel robot is equipped with a self-protection system that triggers an emergency motor stop if a collision occurs.

\section{Method}\label{Section: METHOD}
\subsection{Hybrid hierarchical learning framework}
The assembly process is divided into two network layers: a high-level network and a low-level network, as shown in \fn{Fig_5}. This structure combines the advantages of supervised learning, imitation learning, and reinforcement learning. By using a hierarchical approach, the framework optimizes assembly control, enabling efficient and precise task execution.

\begin{figure*}[h]
	\centering
	\includegraphics[width=160mm]{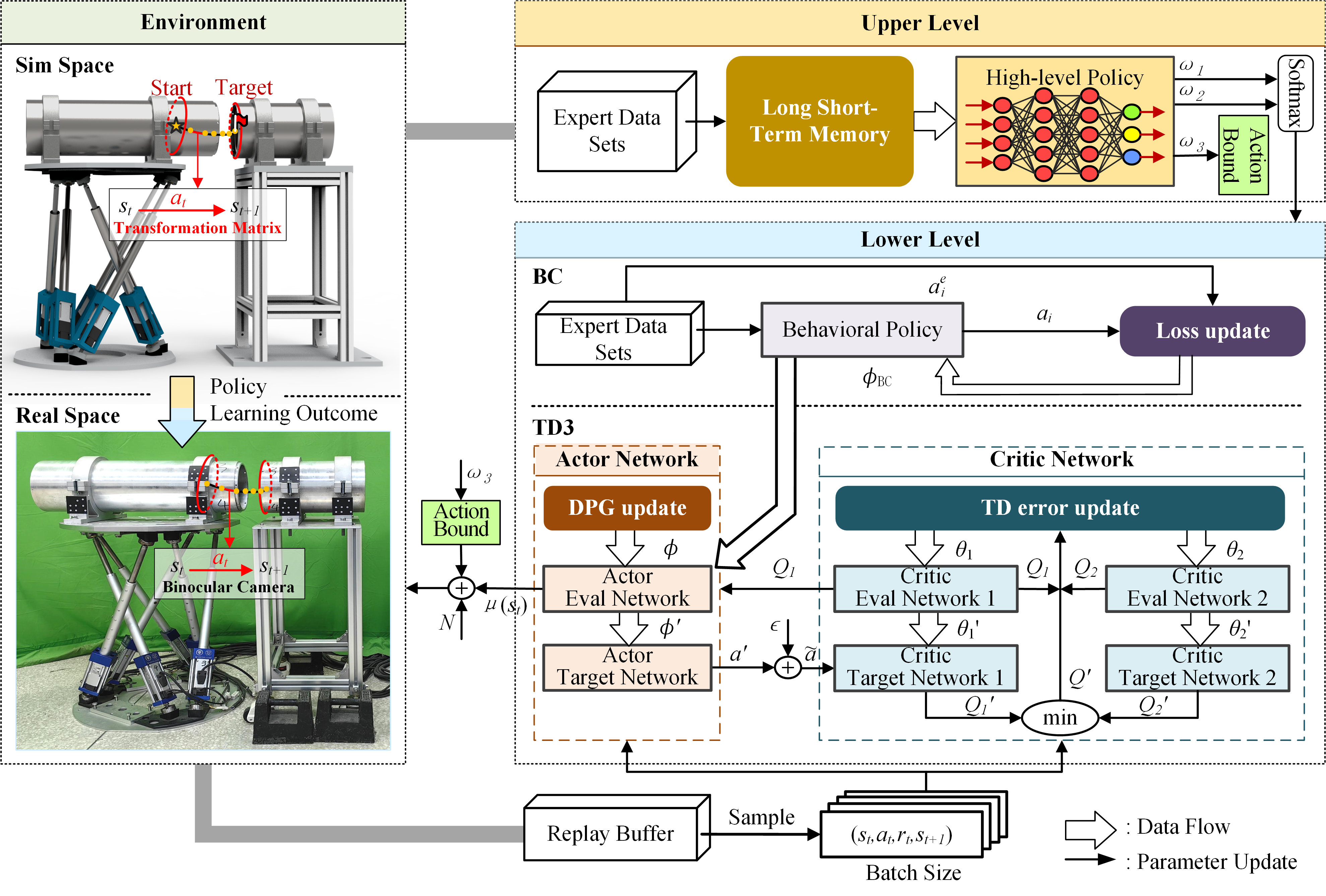}
	\caption{Framework of the hierarchical hybrid algorithm.}
	\label{Fig_5}
\end{figure*}

The high-level network uses LSTM to learn from expert decision-making experiences. This helps in selecting the low-level network and adjusting the step size dynamically during the assembly process. Through its gating mechanisms, LSTM retains important information while discarding irrelevant data, effectively capturing long-term dependencies. In the context of segment assembly, LSTM improves decision accuracy and dynamic adaptability, optimizing the overall process. Its key functions include position and orientation weight selection, dynamic step adjustment weight.

Assembly actions are three degrees of freedom for position and three for orientation. High-level LSTM network generates position selection weight $\omega_1$ and orientation selection weight $\omega_2$. These weights are normalized by softmax to select between the position and orientation sub-networks. The sub-network with the higher probability is activated for adjustment. Typically, the assembly starts with orientation adjustments to achieve the correct docking angle, followed by precise positioning.

High-level LSTM network also generates step-size weights $\omega_3$. The step size is used to dynamically adjust the maximum step limits during assembling. It aims for coarse tuning in the initial stage and fine-tuning in the final stage, ensuring the system maintains efficient and flexible operations under various initial conditions.

The core of LSTM lies in three gate structures: the forget gate, the input gate, and the output gate. When sample data is input into the forget gate, it screens out important information from past memories. Subsequently, the output gate determines how much new information is contained in the current network input. The output value is then calculated using the sigmoid function within the output gate. Finally, the output portion is determined by the tanh function. 

Forget Gate: This gate determines what information to discard from the cell state.
\begin{equation}
f_t=\sigma (W_f\cdot [h_{t-1},s_i]+b_f)
\end{equation}
\noindent where $f_t$ is the forget gate. $\sigma$ is the sigmoid function. $W_f$ and $b_f$ are the weights and biases. $h_{t-1}$ is the previous hidden state. The hidden state conveys past information to subsequent steps and serves as the current output, offering essential context for sequence modeling.

Input Gate: This gate decides what new information to add to the cell state.
\begin{equation}
\begin{aligned}
i_t &= \sigma (W_i \cdot [h_{t-1}, s_i] + b_i)\\
\tilde{c}_t &= \tanh (W_c \cdot [h_{t-1}, s_i] + b_c)
\end{aligned}
\end{equation}

\noindent where $i_t$ is the input gate. $\tilde{c}_t$ is the candidate cell state. $W_i,b_i,W_c,b_c$ are the respective weights and biases.

Output Gate: This gate determines what information to output from the cell state.
\begin{equation}
o_t = \sigma (W_o \cdot [h_{t-1}, s_i] + b_o)
\end{equation}
\noindent where $o_t$ is the output gate. $W_o,b_o$ are the weights and biases.

The LSTM is summarized in Algorithm 1 as follows:

\begin{breakablealgorithm}
  \caption{LSTM algorithm}
  \label{LSTM algorithm}
  \begin{algorithmic}[1]
    \STATE Input: Expert trajectories ${\{(s_i,a_i)\}}_{i=1}^{N}$
    \STATE Randomly initial LSTM parameters $\phi_{LSTM}$
    \STATE $\textbf{for}\; i=0,1,2,...\;\textbf{do}$
    \STATE \hspace{0.5cm} Process trajectory $(s_i,a_i)$ through the LSTM cell
    \STATE \hspace{0.5cm} Compute LSTM gates: ($W$: weight matrices, $b$: bias term) \\
    \hspace{0.5cm} Forget Gate: $f_t=\sigma (W_f\cdot [h_{t-1},s_i]+b_f)$\\
    \hspace{0.5cm} Input Gate: $i_t = \sigma (W_i \cdot [h_{t-1}, s_i] + b_i)$\\
    \hspace{0.5cm} Candidate Cell State: $\tilde{c}_t = \tanh (W_c \cdot [h_{t-1}, s_i] + b_c)$\\
    \hspace{0.5cm} Output Gate: $o_t = \sigma (W_o \cdot [h_{t-1}, s_i] + b_o)$
    \STATE \hspace{0.5cm} Update cell state: $c_t=f_t\cdot c_{t-1}+i_t\cdot\tilde{c}_t$
    \STATE \hspace{0.5cm} Update hidden state: $h_t=o_t\cdot\tanh(c_t)$
    \STATE \hspace{0.5cm} Update parameters $\phi_{LSTM}$ based on loss between predicted actions and expert actions $a_i$:\\
    \hspace{0.5cm} $\phi_{LSTM}\gets\phi_{LSTM}-\beta\cdot\nabla_{\phi_{LSTM}}\mathrm{Loss}(\phi_{LSTM})$\\
    \hspace{0.5cm} $\mathrm{Loss}(\phi_{LSTM})=(2N)^{-1}\textstyle\sum_{i=1}^{N}(h_t-a_i)^2$
    \STATE \textbf{end for}
  \end{algorithmic}
\end{breakablealgorithm}

The low-level network, BC-TD3, controls specific pose-adjusting actions. BC initially learns expert demonstration data to provide a high-quality policy for the actor network. TD3 then optimizes the policy, with BC supplying the initial strategy.

The high-level LSTM and low-level BC-TD3 networks form the hierarchical learning framework. The high-level LSTM generates position, orientation, and step-size adjustment weights. It guides the low-level network. The low-level network uses BC to learn expert strategies and TD3 to further optimize execution, ensuring precise task performance.

\subsection{Imitation learning}
\subsubsection{Behavior cloning}

Imitation learning is a method designed to mimic expert decision-making, with the most straightforward knowledge-guided approach being Behavior Cloning (BC). BC treats expert’s actions as labels and developed the policy by agent, guided by distilled domain knowledge.

Construct an expert training set $D$ consisting of $N$ (state, action) pairs. It is expressed as:
\begin{equation}
D=\{(s_1,a_1^e),(s_2,a_2^e),...,(s_N,a_N^e)\}
\end{equation}
\noindent where $a_i^e$ denotes expert's decision at the $i$-th state $s_i$.

Under the supervised learning paradigm, difference between agent's decisions and expert's decisions at $N$ states is defined as the Behavior Cloning loss:
\begin{equation}
\mathrm{Loss}(\phi_{BC})=\frac{1}{2N}\textstyle\sum_{i=1}^{N}(a_i-a_i^e)^2
\end{equation}
\noindent where $\phi_{BC}$ represents parameters of the agent's policy network. $a_i$ denotes the agent's decision.

The gradient descent of $\mathrm{Loss}(\phi_{BC})$ is used to optimize the policy network parameters $\phi_{BC}$, enabling the agent to achieve decision-making capabilities close to the expert level.

\subsubsection{Imitation learning parameter design}

We distill human expertise to design compact state and action spaces. These spaces preserve spatial cues and retain incremental corrections.

State Space: The coordinate systems obtained from the camera serve as the basis for assembly decision-making and provide spatial relationships between different frames. The state space is defined using real-time visual measurements and transformation matrices to ensure an accurate representation of the system configuration. Specifically, the state describes the relative pose between Segment I and Segment II, determined from encoded points captured by the binocular camera and the transformation matrices in \en{Eq1}. The encoded points establish the coordinate frames of Segment I and Segment II, while the transformation matrices define their relative spatial relationships. Accordingly, the state space is formulated as follows:
\begin{equation}
s_t=[A_{1t},A_{2t},A_{3t},A_{4t},A_{5t},A_{6t},P_{tx},P_{ty},P_{tz},O_{tx},O_{ty},O_{tz}]
\end{equation}
\noindent where $A_{1t},A_{2t},A_{3t},A_{4t},A_{5t},A_{6t}$ are the coordinates of the encoded points in the frame $\{W\}$. $P_{tx},P_{ty},P_{tz}$ are the relative position between frame $\{C_1\}$ and frame $\{C_2\}$. $O_{tx},O_{ty},O_{tz}$ are the relative orientation between them.

Action Space: Assembly actions are divided into position and orientation adjustments. The action space consists of position $a_{\triangle t}$ and orientation $a_{\theta t}$, described as follows:
\begin{equation}
a_{\triangle t}=[\triangle_{tx},\triangle_{ty},\triangle_{tz}],a_{\theta t}=[\theta_{tx},\theta_{ty},\theta_{tz}]
\end{equation}
\noindent where $[\triangle_{tx},\triangle_{ty},\triangle_{tz}]$ are the translational motion of frame $\{M\}$ in $s_t$. $[\theta_{tx},\theta_{ty},\theta_{tz}]$  are the rotational motion of frame $\{M\}$.

The BC is summarized in Algorithm 2 as follows:

\begin{breakablealgorithm}
  \caption{BC algorithm}
  \label{BC algorithm}
  \begin{algorithmic}[1]
    \STATE Input: Expert trajectories ${\{(s_i,a_i)\}}_{i=1}^{N}$
    \STATE Randomly initial policy $\phi_{BC}$
    \STATE \STATE $\textbf{for}\; i=1\;\text{to}\;T\;\textbf{do}$
    \STATE \hspace{0.5cm} Select action with exploration noise $a\thicksim$ 
    \STATE \hspace{0.5cm} Update actor by minimizing the loss: $\mathrm{Loss}(\phi_{BC})=\frac{1}{2N}\textstyle\sum_{i=1}^{N}(a_i-a_i^e)^2$
    \STATE \hspace{0.5cm} Update the policy: $\phi_{BC}\gets\phi_{BC}-\beta\cdot\nabla_{\phi_{BC}}\mathrm{Loss}(\phi_{BC})$
    \STATE \textbf{end for}
  \end{algorithmic}
\end{breakablealgorithm}

\subsection{Reinforcement learning}
\subsubsection{Twin delayed deep deterministic policy gradient}

Reinforcement learning allows assembling policies to be learned through trial-and-error interactions with the environment. The Deep Deterministic Policy Gradient (DDPG) algorithm is the most widely used algorithm for solving continuous control problems. It provides deterministic actions and has better sample efficiency. However, Q-value is usually overestimated and the algorithm is sensitive to noise, affecting robustness in dynamic tasks. TD3 uses dual critic networks, delayed updates, and target policy smoothing to improve stability and performance of high-dimensional continuous control tasks \cite{25}. It has been adapted to various real-world tasks, increasing robustness and success rates in areas such as robotic manipulation and autonomous navigation.

TD3 is a reinforcement learning algorithm based on the Actor-Critic architecture. It consists of two value networks (Critics) and one policy network (Actor). Critic networks update their parameters $\theta _1$ and $\theta _2$ in the direction of minimizing the loss function, thereby maximizing the expected total return. Actor network updates its parameters $\phi $ by following the policy gradient toward the direction of maximum reward, selecting the optimal action to execute. 

Loss functions of the Critic networks are described as follows:
\begin{equation}
\mathrm{Loss}(\theta_k)=\frac{1}{N}\textstyle\sum_{i=1}^{N}(y_i-Q_k(s_i,a_i\mid \theta _k))^2
\end{equation}
\noindent where $N$ represents the number of learning samples selected from experience replay buffer. The target value $y_i$ is defined as:
\begin{equation}
y_i=r_i+\gamma\min_{k=1,2}{Q_k'(s_{i+1},\tilde{a}\mid \theta _k')}
\end{equation}
\noindent where $Q_k,\theta _k$, $(Q_k',\theta _k')$ are the parameters of the current (target) Critic network. $\tilde{a}$ denotes action with added Gaussian noise.

This method ensures a stable target during learning, which improves convergence performance. However, if the error of the Critic network is large, the policy may be discontinuous. To avoid divergence, the Actor network's parameters are updated at a slower rate than those of the Critic network. This slower update process reduces the variance in value function updates, leading to a more robust policy.

Parameters of the Actor network are updated using the backpropagation algorithm. The gradient of loss function expressed as follows:
\begin{equation}
\nabla_\phi J(\phi)\approx \frac{1}{N}\sum_{i=1}^{N}[\nabla_aQ_1(s_i,a\mid\theta^Q)\mid_{a=\mu (s_i)}\cdot\nabla_\phi\mu(s_i)]
\end{equation}
\noindent where $\nabla$ represents gradient, $\mu$ and $\theta$ are parameters of the current Actor network.

To avoid overfitting, Q-value is smoothed by adding noise that follows a truncated normal distribution to each action. The target update is modified as:
\begin{equation}
\tilde{a}\gets a'+\varepsilon, \varepsilon \thicksim \mathrm{clip}(N(0,\sigma),-c,c),c>0
\end{equation}
\noindent where $a'=\mu'(s_{i+1})$. $\varepsilon$ represents the noise following a truncated normal distribution. $\sigma$ is the variance. $c$ is the clipping threshold. “clip” is a method to limit the range of noise added to actions between $-c$ and $c$, ensuring stability during training.

\subsubsection{Reinforcement learning parameter design}

Execution of assembly task is modeled as a Markov Decision Process. At each time step $t$, a state $s_t$ is obtained, an assembly action $a_t$ is chosen, and a reward $r_t$ is received to evaluate the action-state pair. The learned policy is improved by maximizing the return $R_t$.
\begin{equation}
R_t=r_t+\gamma r_{t+1}+\gamma^2 r_{t+2}+...\gamma^{T-t} r_{T}=r_t+\gamma R_{t+1}
\end{equation}
\noindent where $r_t$ depends on the performance of action $a_t$ in state $s_t$. T is the final time step. $\gamma\in [0,1]$ is the discount factor, which decreases the value of rewards received later.

Global Reward Function Design: At time $t$, the agent performs an action $a_t$, and the binocular camara provides the state $s_t$. Relative position and orientation deviations between Segment I and II are calculated. If assembly accuracy is not met, the process moves to the next step. Upon successful assembly or failure within the specified steps, the system receives reward feedback, and the cycle ends. The reward function is as:
\begin{equation}
R_g=\left\{\begin{matrix}
r_\mathrm{su},\  \mathrm{if\ success}
\\ r_\mathrm{fa},\  \mathrm{if\ failure}
\\ 0,\  \mathrm{otherwise}
\end{matrix}\right.
\end{equation}
\noindent where $r_{su}$ is the reward value obtained upon successful assembly. $r_{fa}$ is the penalty value for failure to complete the assembly within the specified number of steps, for exceeding motion limits or causing collisions between components. In other cases, the feedback for the agent is 0. In each training round, the target reward $R_g$ is awarded only once and concludes the training round. Set $r_\mathrm{su}=50$ and $r_\mathrm{fa}=-50$.

To enhance the exploration efficiency of the agent and encourage the robot to complete assembly with shortest possible path, a penalty term of -0.01 is added at each step.

Local Reward Function Design: For the two agents controlling position and orientation separately, independent reward functions are set according to their task characteristics.

Position-Based Reward Function:
\begin{equation}
r_P=\left\{\begin{matrix}
1-[(P_{tx}^2+P_{ty}^2+P_{tz}^2)^{1/2}/P^\mathrm{init}]
\\ 1-[\begin{vmatrix}P_{tz}\end{vmatrix}/\begin{vmatrix}P_z^\mathrm{init}\end{vmatrix}]
\end{matrix}\right.
\end{equation}
\noindent where $P^\mathrm{init}$ is the distance between the Segment I and II. $\begin{vmatrix}P_z^\mathrm{init}\end{vmatrix}$ is the axial distance.

Orientation-Based Reward Function:
\begin{equation}
r_O=1-[(O_{tx}^2+O_{ty}^2+O_{tz}^2)^{1/2}/O^\mathrm{init}]
\end{equation}
\noindent where $O^\mathrm{init}$ is the dot product of the angles between two segments.

\subsubsection{Reward function constraints}

The reward function incorporates position and orientation penalty terms to accelerate training convergence and constrain robot's behavior. Incorrect positions and orientations during training can be avoided.
\begin{equation}
r_\mathrm{err}=-5,\mathrm{if\ count}P>20\ \mathrm{or}\ \mathrm{count}O>20
\end{equation}
\noindent where $\mathrm{count}P$ is the count accumulated when $\begin{Vmatrix}P_t\end{Vmatrix}>\begin{Vmatrix}P_{t-1}\end{Vmatrix}$. $\mathrm{count}O$ is the count accumulated when $\begin{Vmatrix}O_t\end{Vmatrix}>\begin{Vmatrix}O_{t-1}\end{Vmatrix}$.

The final reward function is a weighted sum of global and local rewards:
\begin{equation}
R=\lambda _gR_g+\sum_{t=1}^{T}(\lambda_Pr_{P,t}+\lambda_Or_{O,t}+\lambda_\mathrm{err}r_{\mathrm{err},t})
\end{equation}
\noindent where $R_g$ is the global reward for the entire episode. $r_{P,t},r_{O,t},r_{\mathrm{err},t}$ are local rewards at each time step. $\lambda_g,\lambda_P,\lambda_O,\lambda_\mathrm{err}$  determine the relative importance of each reward component and balance their contributions. Local rewards are summed over all time steps within the episode, whereas the global reward is assigned once per episode. The global reward evaluates the overall success of the task. The local rewards refine position and orientation accuracy. These penalty constraints ensure stable training and precise assembly.

The TD3 is summarized in Algorithm 3 as follows:

\begin{breakablealgorithm}
  \caption{TD3 algorithm}
  \label{TD3 algorithm}
  \begin{algorithmic}[1]
    \STATE Initialize critic networks $Q_1$, $Q_2$ with random parameters, and initialize actor network $\mu$ with the result of BC $\phi_{BC}$
    \STATE Initialize target networks $\theta_1'\gets\theta_1,\ \theta_2'\gets\theta_2,\ \phi'\gets\phi$
    \STATE Initialize replay buffer
    \STATE $\textbf{for}\; t=1\ \textbf{to}\ T\ \textbf{do}$
    \STATE \hspace{0.5cm} Select action with exploration noise $a\thicksim\mu(s_i)+\varepsilon,\ \varepsilon \thicksim N(0,\sigma )$ 
    \STATE \hspace{0.5cm} Execute action $a$, and observe reward $r$ and next state $s'$
    \STATE \hspace{0.5cm} Store transition tuple $(s,a,r,s')$ in replay buffer
    \STATE \hspace{0.5cm} Sample mini-batch of $N$ transitions $(s,a,r,s')$ from replay buffer 
    \STATE \hspace{0.5cm} Compute target action $\tilde{a}\gets \mu'(s')+\varepsilon', \varepsilon' \thicksim \mathrm{clip}(N(0,\sigma),-c,c),c>0$
    \STATE \hspace{0.5cm} Compute target Q-value $y=r+\gamma\min_{k=1,2}{Q_k'(s',\tilde{a}\mid \theta _k')}$
    \STATE \hspace{0.5cm} Update critics by minimizing the loss: $\frac{1}{N}\textstyle\sum(y-Q_k(s,a\mid \theta _k))^2,\ k=1,2$
    \STATE \hspace{0.5cm} $\textbf{if}\; t\ \textbf{mod}\ d\ \textbf{then}$
    \STATE \hspace{1cm} Update the actor by the deterministic policy gradient: $\nabla_\phi J(\phi)\approx \frac{1}{N}\sum[\nabla_aQ_1(s,a\mid\theta^Q)\mid_{a=\mu (s)}\cdot\nabla_\phi\mu(s)]$
    \STATE \hspace{1cm} Update target networks: $\theta_k'\gets\omega\theta_k+(1-\omega)\theta_k',k=1,2,\ \phi'\gets\omega\phi+(1-\omega)\phi'$
    \STATE \textbf{end for}
  \end{algorithmic}
\end{breakablealgorithm}

\section{Simulation}\label{Section: SIMULATION}
\subsection{Simulation Model Setup}

The key to building the simulation model is accurately simulating the effects of actions on the system state to ensure alignment with real-world behavior. The model should prioritize functional integration for future expansion and applications, as shown in \fn{Fig_6}.

\begin{figure}[h]
    \centering
    \includegraphics[width=0.6\columnwidth]{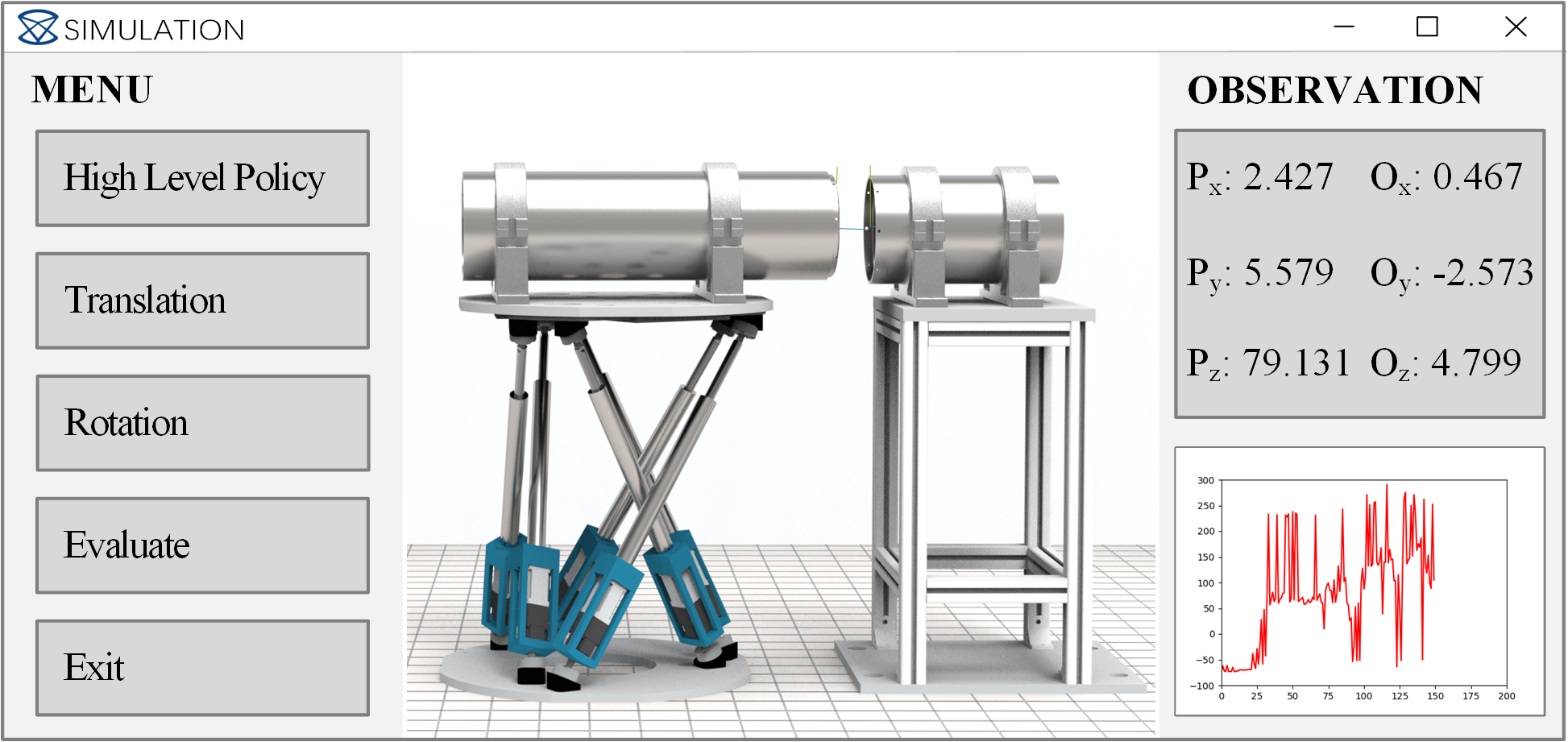}
    \caption{Components of the simulation.}
    \label{Fig_6}
\end{figure}

Initialization: Set transformation matrices for segment I, segment II, and the moving platform. Initialize positions of six encoded points and define the rod lengths of the parallel robot.

Pose Determination: Construct frame $\{C_1\}$ and frame $\{C_2\}$ using encoded points. Establish frame $\{M\}$, $\{F\}$ and $\{P\}$ for each segment through transformation matrices. Determine the robot’s coordinate frame $\{O\}$ using forward kinematics of the parallel robot, based on frame $\{P\}$ and rod lengths.

Control Strategy and Execution: Obtain the current state from the pose, generate action commands using the policy network, calculate new positions of encoded points with control actions, update the state, determine rod lengths through inverse kinematics, and send them to actuators.

Reward Acquisition: Calculate the reward for the current round, providing feedback to the control network to optimize control strategies during iterations.

To ensure stable training and optimal performance, the networks are trained with appropriately chosen hyperparameters. The LSTM model is trained for 10000 epochs using the Adam optimizer, while the BC model is trained for 10000 epochs with the SGD optimizer. Both the Actor and Critic networks in TD3 use the Adam optimizer. The primary training hyperparameters are determined through testing and are summarized in \tn{table1}.

\begin{table}[H]\footnotesize 
  \center
  \caption{Training hyperparameters}
  \label{table1}
  \resizebox{0.4\linewidth}{!}{ 
    \begin{tabular}{cc}
      \toprule 
      \textbf{Parameters}              & \textbf{Value} \\ 
      \midrule
      Mini batch size & 32   \\
      LSTM learning rate & 1e-3   \\
      BC learning rate   & 1e-3   \\
      TD3 Actor/Critic learning rate  & 3e-4/3e-3  \\ 
      TD3 exploration noise clip $\varepsilon$   & 0.1   \\
      TD3 delayed policy updates $d$   & 2  \\
      Target network update rate ($\tau$) & 5e-3 \\
      Experience replay buffer size & 1e6 \\
      Reward discount factor $\gamma$ & 0.98\\
      Reward weight $\lambda_g,\lambda_P,\lambda_O,\lambda_\mathrm{err}$ & 1,0.3,0.3,0.4\\
      \bottomrule
    \end{tabular}
  }
\end{table}

\subsection{Algorithms training with simulation model}
\subsubsection{Learning efficiency}

To verify the advantages of the hybrid layered algorithm, we compared our algorithm with TD3, Normalized Advantage Functions (NAF) and DDPG. As shown in \fn{Fig_7}\blue{a}, reward values of different algorithms vary with the number of training steps. Our algorithm demonstrates significantly higher learning efficiency at the early stages, with the reward curve rising rapidly and achieving successful assembly within 500 episodes. In contrast, traditional single-agent reinforcement learning algorithms require over 1000 episodes to reach a similar reward level and exhibit larger fluctuations during training.

\begin{figure}[h]
    \centering
    \includegraphics[width=0.6\columnwidth]{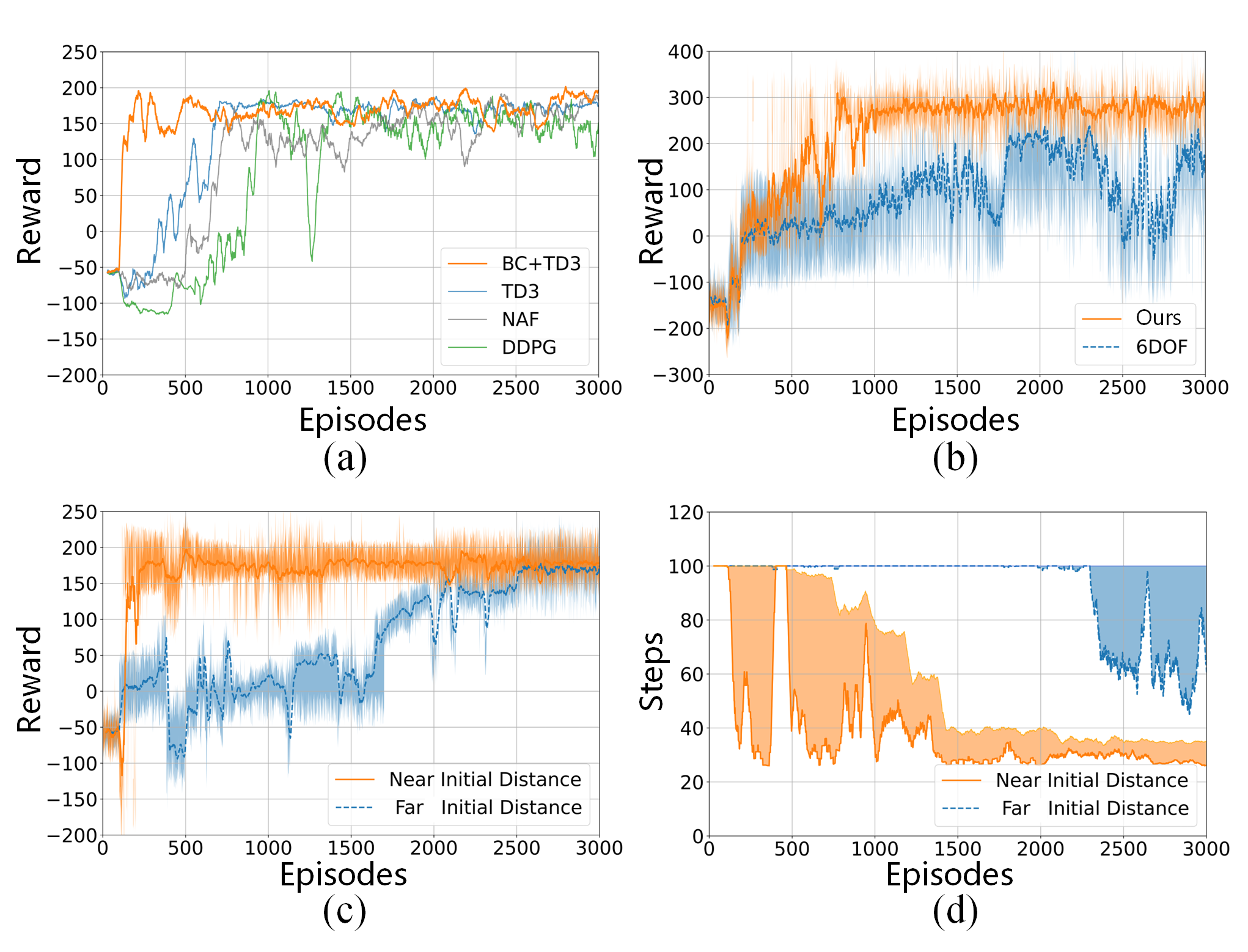}
    \caption{Results of our method and comparative methods in the training stage. (a) Reward curves of comparative simulations. (b) Reward curves for different learning strategies. (c) Reward curves for different initial distances. (d) Steps curves for different initial distances.}
    \label{Fig_7}
\end{figure}

\fn{Fig_7}\blue{b} shows the reward curves under different learning strategies. Our algorithm adopts a 3-DoF rotational model and then a 3-DoF translational model. For comparison, the 6-DoF model is applied without separating rotations and translations. It is found that our learning strategy leads to higher learning efficiency. The 6-DoF model struggles to achieve successful assembly within maximum number of episodes, which is due to the highly non-linear nature of the learning problem.

\fn{Fig_7}\blue{c} illustrates the reward curves with different initial conditions. A closer initial distance results in higher rewards earlier in training, while a farther initial distance yields the opposite effect. This is because high-precision assembly depends primarily on the final docking phase, and with a closer initial distance, the agent can quickly explore various docking scenarios and identify effective assembly routes. \fn{Fig_7}\blue{d} shows the number of steps required for different initial distances, further confirming this observation.

\subsubsection{Assembly efficiency}

\fn{Fig_8}\blue{a} shows an inverse exponential relationship between the initial distance and the number of steps required for assembly. \fn{Fig_8}\blue{b} highlights that step length decreases as the number of steps increases. The model with a dynamic step adaptation strategy allows larger steps for faster movement when the segment is far from the target, and smaller steps for precise adaptation as it approaches. This leads to high efficiency while maintaining high accuracy in the long-distance assembly. In fact, such assembling strategy matches the manual docking process, which has been proven to be effective. 

\begin{figure}[h]
    \centering
    \includegraphics[width=0.6\columnwidth]{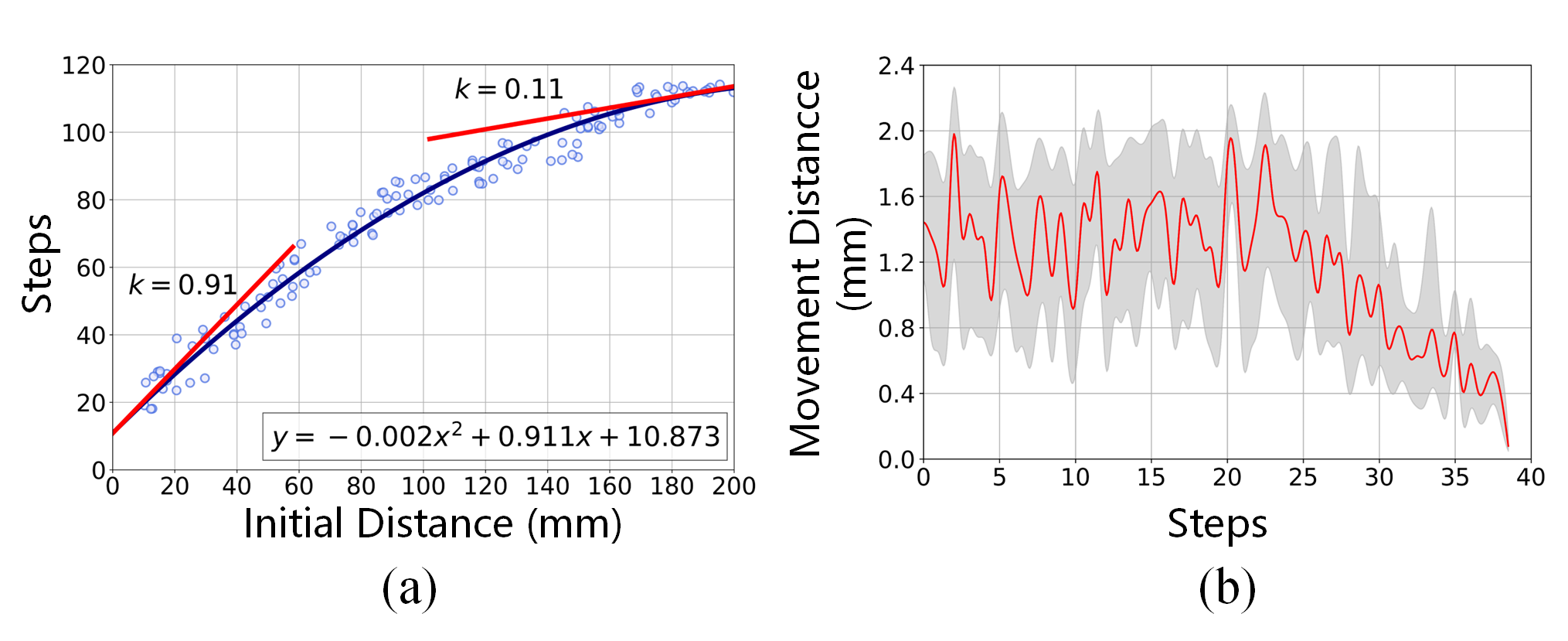}
    \caption{Assembly efficiency with initial distance variation: (a) Effect of initial distance on the number of steps. (b) Relationship between steps and movement distance (initial distance: 50mm).}
    \label{Fig_8}
\end{figure}

\fn{Fig_9} visualizes assembly simulation actions, showing how adjustments change at different stages during leaning. It indicates that hieratical learning strategy for dynamic step adjustment is more effective and efficient compared to the RL learning algorithm alone.

\begin{figure}[h]
    \centering
    \includegraphics[width=0.6\columnwidth]{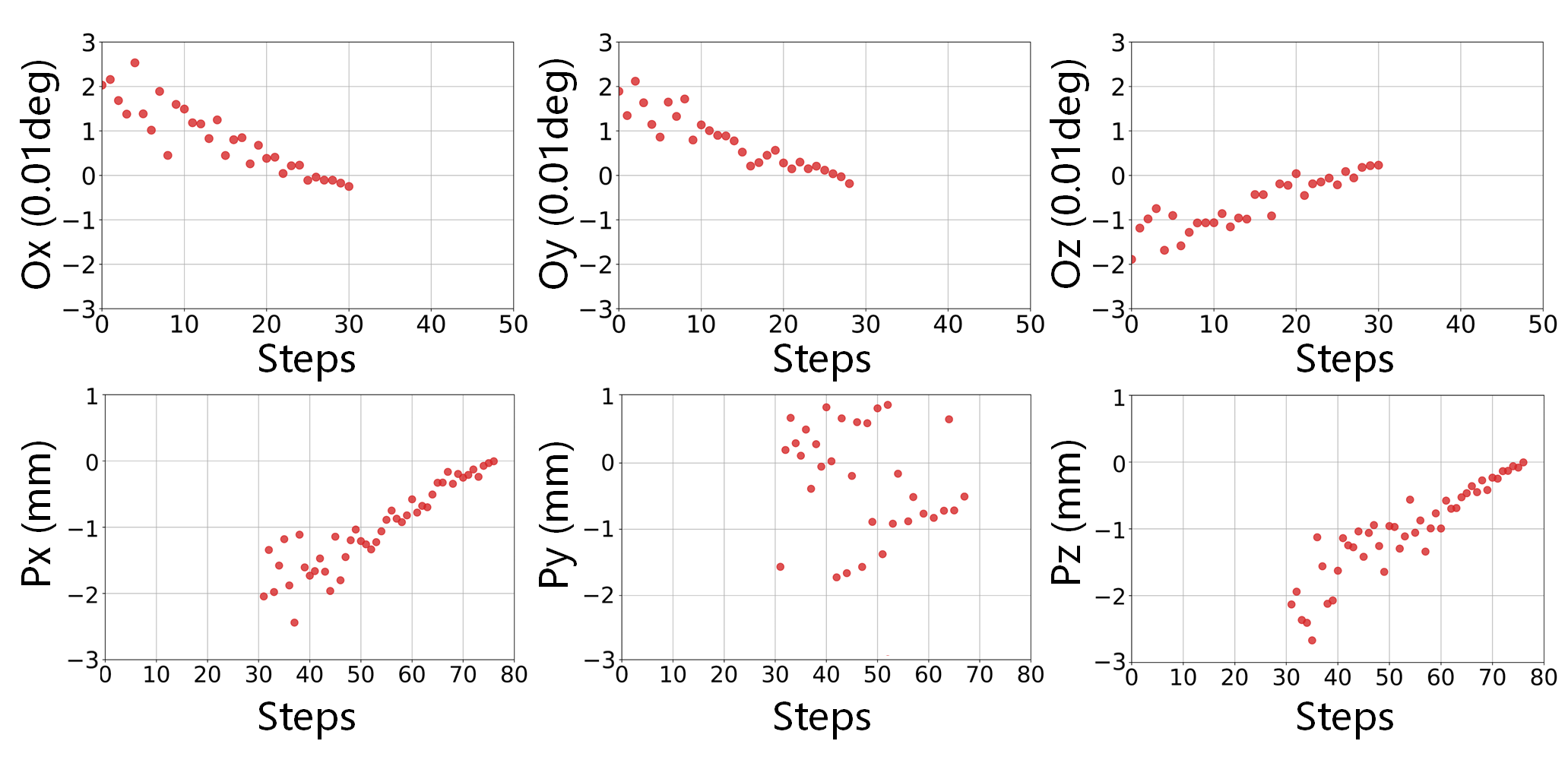}
    \caption{Actions in the segment assembly simulation (initial distance: 100mm).}
    \label{Fig_9}
\end{figure}

\subsubsection{Model adaptability}

We evaluated the algorithm's adaptability to initial errors by different starting conditions. Random errors were introduced at each training step, with position errors of 1, 3, and 5 mm and orientation errors of 1, 2, and 3 degree. Testing conditions included initial poses set with positions of 10, 30, and 50 mm and orientations of 3, 6, and 9 degree. 

\begin{table}[H]
\footnotesize
\centering
\caption{Assembly success rate for pure position and orientation error}
\label{table2}
\resizebox{0.4\linewidth}{!}{
    \begin{tabular}{lcccc}
        \toprule
        Success Rate & \multicolumn{4}{c}{Initial Position and Orientation} \\
        \midrule
        &  & 10 mm & 30 mm & 50 mm \\
        \cmidrule(lr){2-5}
        \multirow{3}{*}{Position}  
        & 1 mm & 1.0 & 1.0 & 0.983 \\
        & 3 mm & 1.0 & 0.956 & 0.869 \\
        & 5 mm & 0.946 & 0.682 & 0.606 \\
        \midrule
        &  & 3 deg & 6 deg & 9 deg \\
        \cmidrule(lr){2-5}
        \multirow{3}{*}{Orientation}  
        & 1 deg & 0.759 & 0.624 & 0.487 \\
        & 2 deg & 1.0 & 0.854 & 0.784 \\
        & 3 deg & 0.928 & 0.822 & 0.573 \\
        \bottomrule
    \end{tabular}
    }
\end{table}

\tn{table2} shows success rates for pure position and orientation errors. For position errors, training with a 1 mm error yielded the highest success rate (above 0.9). As the error increased to 3 mm and 5 mm, the success rate decreased to below 0.7 at 5 mm. The alignment challenge is the final assembly. Smaller position errors allowed more alignment attempts, creating an optimized strategy. In testing, the success rate further declined as the initial position error increased from 10 mm to 50 mm, showing reduced adaptability with larger distances. For orientation errors, a lower success rate of around 0.6 was observed with a 1 degree error. Increasing to 2 degrees raised the success rate above 0.7, but 3 degrees caused a decline, suggesting that 2 degrees is optimal. Larger angles made it more difficult to develop an effective strategy within the fixed episodes. Similar trends were seen with decreasing success rates as the initial orientation angle increased from 3 to 9 degrees.

\begin{table}[H]
	\footnotesize 
	\centering
	\caption{Assembly success rate for combined errors}
	\label{table3}
	\resizebox{0.6\linewidth}{!}{
		\begin{tabular}{lcccc}
			\toprule
			\multirow{2}{*}{\raisebox{2ex}{Success Rate}} & \multicolumn{4}{c}{Initial Position and Orientation} \\
			\midrule
			&  & 3 deg & 6 deg & 9 deg \\
			\cmidrule(lr){2-5}
			\multirow{3}{*}{Position(1mm),Orientation(2deg)} 
			& 10 mm & 0.989 & 0.851 & 0.769 \\
			& 30 mm & 0.928 & 0.795 & 0.613 \\
			& 50 mm & 0.916 & 0.651 & 0.598 \\
			\bottomrule
		\end{tabular}
	}
\end{table}

\tn{table3} shows success rates when both position and orientation errors were present. For training, we selected optimal conditions, fixing the position error at 1 mm and the orientation error at 2 degrees. Under these dual-error conditions, the overall success rate decreased. The highest success rate of 0.989 was achieved with a 10 mm position and a 3 degree orientation. As the initial position and orientation increased, the success rate gradually decreased, reaching the lowest value of 0.598 for a 50 mm position and a 9 degree orientation.

Data from \tn{table2} and \tn{table3} clearly indicate that the algorithm’s success rate is influenced by training error levels under pure position or orientation errors. As testing errors increase, the success rate decreases. Under combined position and orientation errors, the algorithm demonstrates slightly weaker adaptability but maintains a high level of stability.

\subsubsection{Model robustnes}

In the simulation environment, the robot's state $s_t$ is $[A_{1t},A_{2t},A_{3t},A_{4t},A_{5t},A_{6t},P_{tx},P_{ty},P_{tz},O_{tx},O_{ty},O_{tz}]$. To simulate the uncertainties in the real environment, Gaussian noise is added to the state observations, and the state $s_t$ becomes:
\begin{equation}
s_t'=s_t+N(0,\sigma^2)
\end{equation}
\noindent where $N(0,\sigma^2)$ represents a Gaussian distribution with a mean of 0 and a variance of $\sigma^2$.

\begin{table}[H]\footnotesize 
  \center 
  \caption{Results of the success rate with different noise levels [$\sigma$]}
  \label{table4}
  \resizebox{0.65\linewidth}{!}{ 
    \begin{tabular}{cccccc}
        \toprule
        Episodes range & 0-400 & 400-800 & 800-1200 & 1200-1600 & 1600-2000 \\
        \midrule
        0 & 0.698 & 0.977 & 0.983 & 0.991 & 0.987 \\
        0.02 & 0.385 & 0.706 & 0.814 & 0.918 & 0.974 \\
        0.05 & 0.197 & 0.433 & 0.767 & 0.875 & 0.867 \\
        0.1 & 0.156 & 0.379 & 0.405 & 0.574 & 0.620 \\
        0.2 & 0.099 & 0.216 & 0.261 & 0.315 & 0.328 \\
        \bottomrule
    \end{tabular}
  }
\end{table}

We added Gaussian noise with a mean of zero and a standard deviation that gradually increased while keeping the initial position and orientation unchanged. Results in \tn{table4} show the assembly success rate decreases as noise level increases. When the number of training episodes is sufficient, the success rate remains relatively high. This shows our model has good robustness and works well transferring from simulation to real-world applications. The assembly gap is 0.1mm. Successful assembly is difficult if the measurement error exceeds the assembly gap in the traditional assembly methods based on general trajectory planning. Comparatively, our method still reaches a good success rate when the noise level is 0.1 or even 0.2. The fact that our model maintains a reasonable success rate under such conditions further proves its robustness.

\section{Experiments and analysis}\label{Section: EXPERIMENTS AND ANALYSIS}

\begin{figure}[h]
    \centering
    \includegraphics[width=0.5\columnwidth]{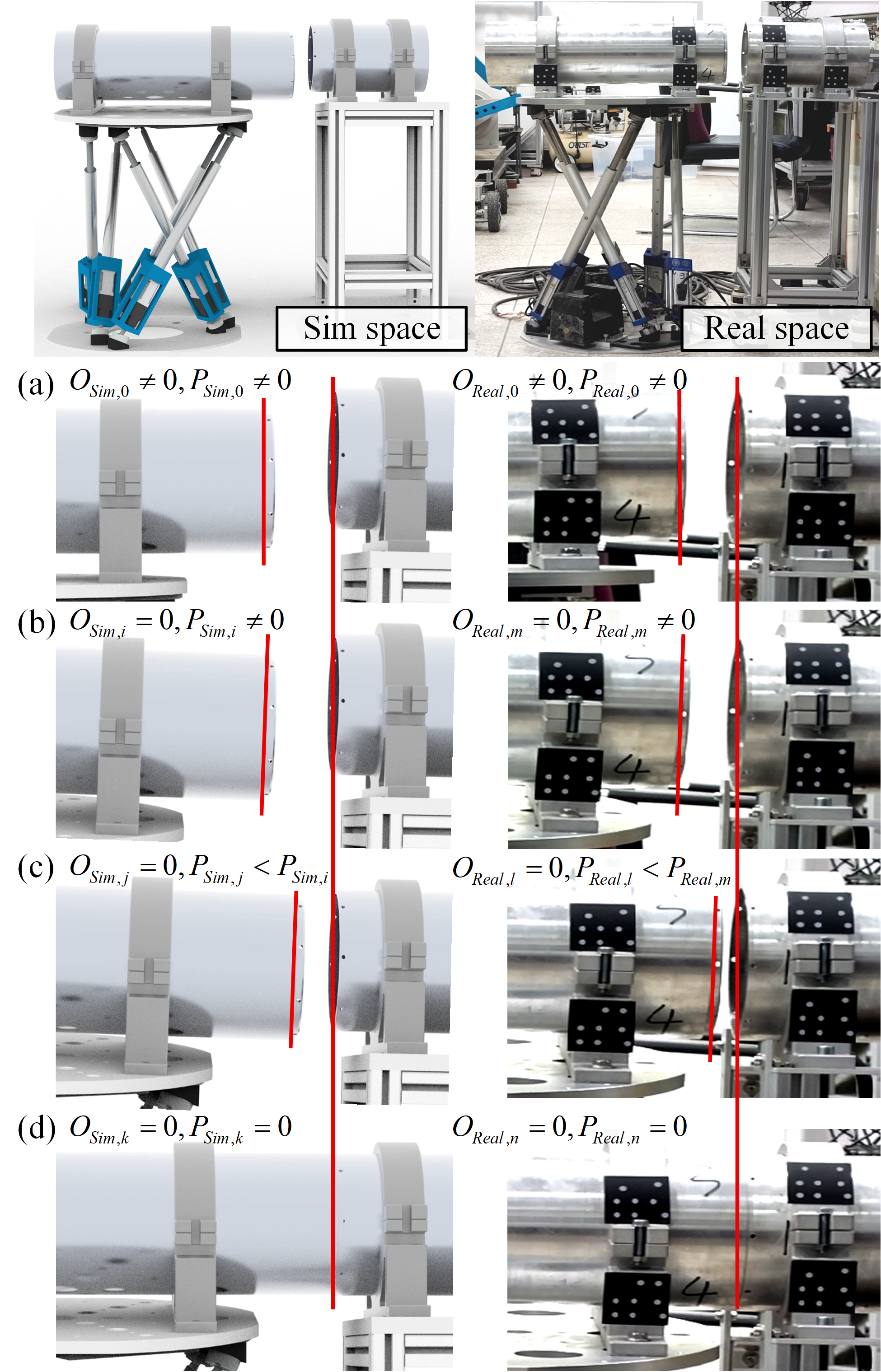}
    \caption{Assembly comparison between simulation and reality: (a) Initial state. (b) Rotation-dominated. (c) Translation-dominated. (b) Complete assembly movement.}
    \label{Fig_10}
\end{figure}

Simulation environment is precisely built according to the robotic system, leading to high accuracy of simulation to reality mapping. During the transfer process, coordinate mapping is used to align the coordinates of encoded points in the real environment with those in the simulation environment. This involves aligning the coordinate systems of the simulation and real environments to ensure consistency between the source domain (simulation) and target domain (reality). The effectiveness of this mapping is demonstrated in \fn{Fig_10}.

The effectiveness of this mapping is demonstrated in \fn{Fig_10}. Identical initial poses are set in sim and real. Snapshots are taken at close time instants. The encoded-point paths reveal a Knowledge-Guided strategy. The robot first rotates the segment until the orientation error $O$ equals zero. It then translates until the hole offset $P$ equals zero. Sim and real motions follow this learned assembly sequence.

\subsection{Trajectory smoothness}

Based on pre-trained simulation models, each algorithm underwent an additional 100 training iterations in a real environment to ensure practical effectiveness. To evaluate performance, we conducted 20 experiments for each algorithm under various initial poses. \fn{Fig_11}\blue{a} illustrates the motion trajectories of different algorithms during the assembly. As shown, our algorithm demonstrates smoother motion with less fluctuation in the trajectory, outperforming the methods in \cite{26} and \cite{27} in terms of stability and continuity. In \fn{Fig_11}\blue{b}, a close-up view of the trajectories reveals the detailed performance of each algorithm at key positions. Our algorithm exhibits significant advantages in smoothness. Also, it requires fewer steps to reach the same assembly depth.

\begin{figure}[h]
    \centering
    \includegraphics[width=0.6\columnwidth]{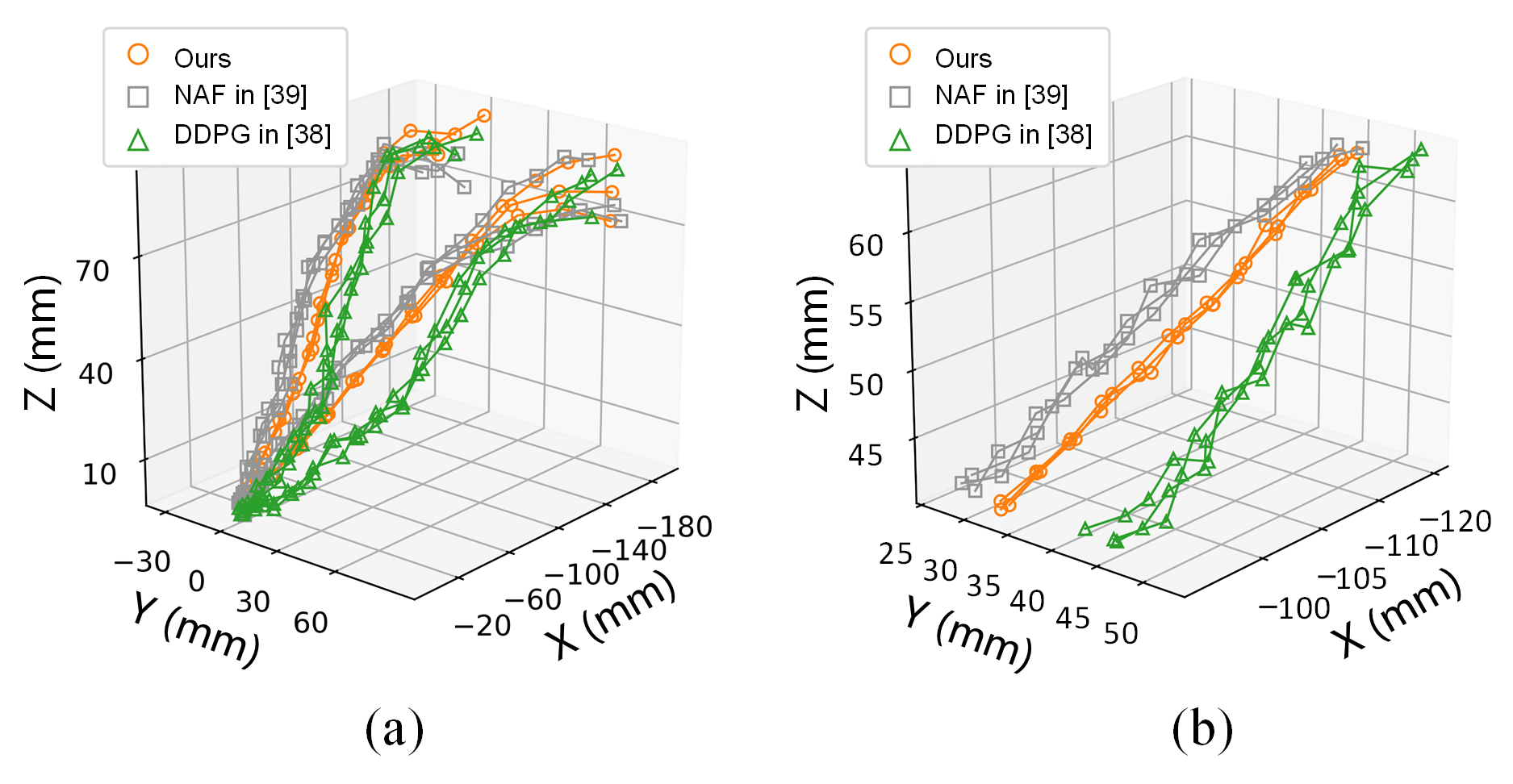}
    \caption{Trajectories of Segment I during the assembly experiment: (a) Complete trajectories. (b) Partial trajectories (x: -120~-90mm, y: 20~60mm, z: 40~70mm).}
    \label{Fig_11}
\end{figure}

Further analysis is shown in \fn{Fig_12}, where the orientation (Ox, Oy, Oz) and position (Px, Py, Pz) adjustments of Segment I during assembly experiment are compared with different algorithms. It presents the trajectory of pose and position changes for each algorithm. The results show that our algorithm outperforms methods in \cite{26} and \cite{27} in pose adjustment and precise position control. Particularly in the final stage of assembly, our algorithm stabilizes more quickly and maintains smaller errors across all dimensions. A comparison of the required adjustment steps shows that our algorithm requires significantly fewer steps, achieving a stable state more rapidly. This indicates that our method optimizes the assembly path while substantially reducing unnecessary adjustment steps.

\begin{figure}[h]
    \centering
    \includegraphics[width=0.6\columnwidth]{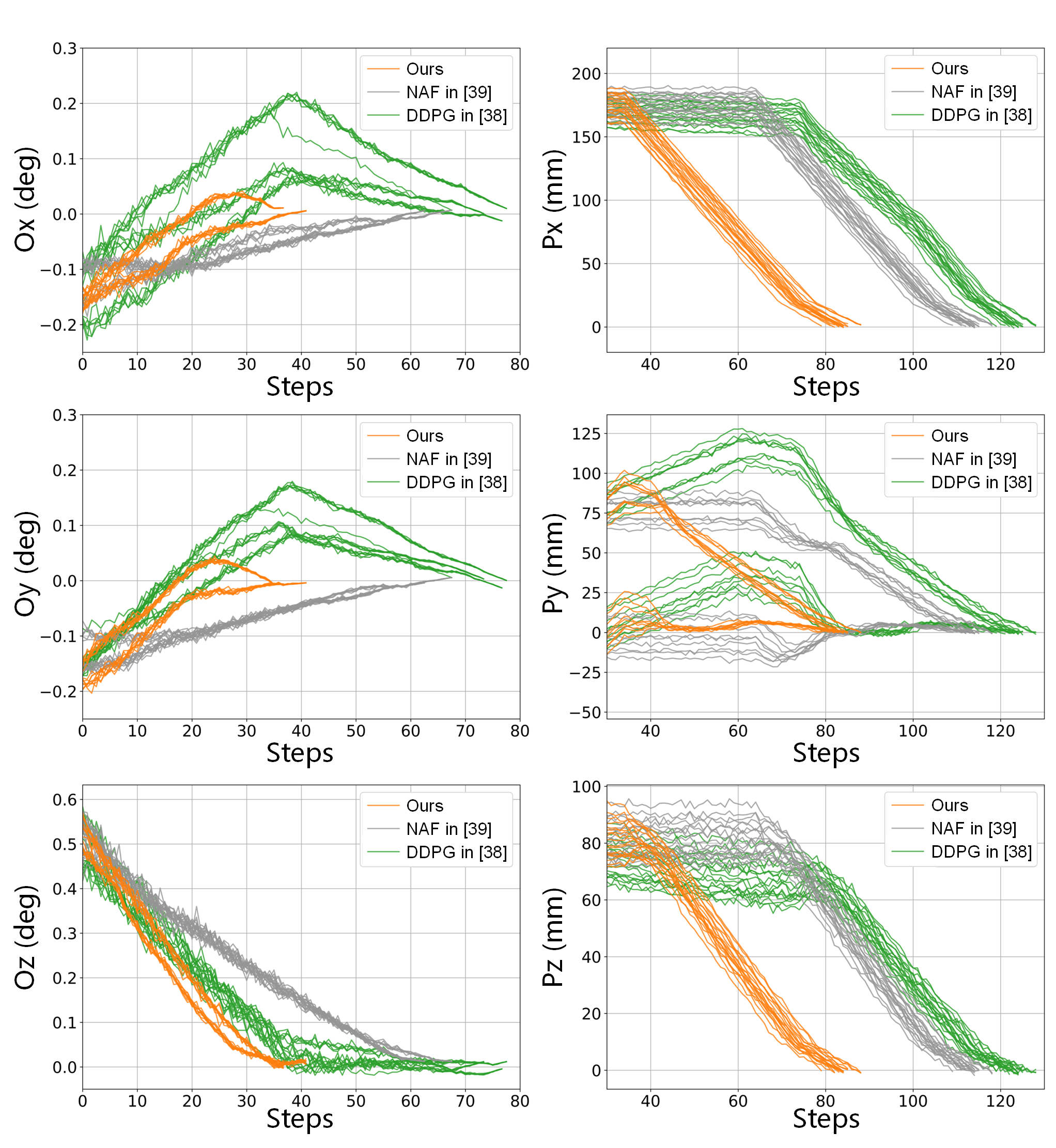}
    \caption{Comparison of pose and position for different algorithms during the assembly experiment of Segment I.}
    \label{Fig_12}
\end{figure}

\subsection{Transferability}

In both simulation and real-world settings, experiments were conducted using the trained algorithm to execute assembly task under consistent initial conditions, with an initial position of 50 mm and an initial orientation of 1 deg. These algorithms were trained for the same number of iterations in the simulation environment and then transferred to the real environment for additional training with the same number of iterations. Meanwhile, we compared Heuristic Methods and Learning from Demonstration (LFD).

For heuristic methods, comparative experiments were carried out with two schemes: visual servoing and spiral search. The visual servoing scheme adopted the method in \cite{22}, and it employed the assembly process of pose adjustment at the mooring point combined with translational docking. It conducts multiple image captures at the mooring point to acquire the actual pose of the robot's moving platform, and adjusts the robot's pose at the mooring point according to the pose of the fixed cylindrical segment to complete the assembly. The spiral search scheme adopted the method in \cite{add13}, integrating two in-hand cameras and a force-torque sensor to compensate for positional uncertainty and implementing a sequence of learning-based visual servoing, spiral search and impedance control for the peg-in-hole assembly task.

We perform coarse positioning via binocular visual servoing to guide Segment I to the vicinity of Segment II. The spiral search is implemented on the $xy$-plane of frame $\{M\}$. A reference coordinate system for spiral search, denoted as frame $\{S\}$, is established with the same orientation as frame $\{M\}$, and its origin is offset by $r$ from the initial peg position. The discrete spiral search path is defined as:
\begin{equation}
	x_s = r \cos\theta, \quad y_s = r \sin\theta
\end{equation}
\noindent where the initial values of $\theta$ and $r$ are both 0. The angle $\theta$ increases by $\delta\theta$ at each timestep, while the radius $r$ increases by $\delta r$ after every full rotation.

Since the force sensor is not employed in this work, the force-based termination condition in the method of \cite{add13} is replaced with a vision-based detection method. The robot advances along the $z$ direction of frame $\{S\}$ by a small step after each discrete spiral movement in the $xy$-plane. If no collision is detected, it continues to move forward along the $z$ direction, while if a collision is detected, it retreats and resumes the spiral movement.

For LFD, comparative experiments were conducted with two schemes: BC and Dynamic Movement Primitives (DMPs). The BC scheme trained the network directly using the collected training datase.  DMPs stem from the motor control of biological systems and serve as a rigorous mathematical formulation of motion primitives as stable nonlinear dynamical systems \cite{add14}.The DMPs scheme adopted the method in \cite{add15}, which proposes a demonstration-trajectory adaptation-enhanced DMP (DA-DMP) based approach for multi-procedure robotic assembly, constructing a 7-dimensional Cartesian DMP with 3D position and 4D quaternion orientation trajectories.

The position information $P_{x},P_{y},P_{z},O_{x},O_{y},O_{z}$ in the state $s$ field of the expert dataset is extracted, and the original Euler angle attitude information is converted to quaternion form $q_w, q_x, q_y, q_z$, which are integrated into a 7-dimensional pose time series with 3D position and 4D attitude. The DA-DMP method is rooted in a point attractive  system, which aligns expert trajectories to new task configurations. Its fundamental dynamic equation is expressed as:
\begin{equation}
\tau \ddot{y} = \alpha \left( \beta (g - y) - \dot{y} \right) + f
\end{equation}
\noindent where $\tau$ is the time scaling factor. $y$ is the 7-dimensional pose vector. $g$ is the control target. $\alpha$ and $\beta$ are gain coefficients analogous to the proportional-derivative controller gains. $f$ is the nonlinear function.

\begin{table}[H]
	\footnotesize 
	\centering
	\caption{Comparison results of the average steps, average reward and success rate}
	\label{table5}
	\resizebox{\linewidth}{!}{
		\begin{tabular}{lccccccc}
			\toprule
			Method & \multicolumn{3}{c}{Sim. Environment} & \multicolumn{3}{c}{Real Environment} \\
			\cmidrule(lr){2-4} \cmidrule(lr){5-7}
			& Avg. Steps & Avg. Reward & Success Rate & Avg. Steps & Avg. Reward & Success Rate \\
			\midrule
			\multicolumn{7}{l}{\textit{Heuristic Methods}} \\
			Visual Servoing \cite{22} & - & - & 1.00 & - & - & 0.75 \\
			Spiral Search \cite{add13} & 209 & - & 0.97 & 283 & - & 0.60 \\
			\midrule
			\multicolumn{7}{l}{\textit{Learning from Demonstration}} \\
			Behavior Cloning & 145 & 75.1 & 0.73 & 183 & 40.7 & 0.55 \\
			DA-DMP \cite{add15} & 114 & 93.7 & 0.93 & 149 & 71.5 & 0.70 \\
			\midrule
            Propose method & 47 & 153.7 & 0.99 & 48 & 149.9 & 1.00 \\
            Method in \cite{27} & 62 & 120.4 & 0.86 & 69 & 100.1 & 0.90 \\
            Method in \cite{26} & 57 & 135.1 & 0.91 & 60 & 123.7 & 1.00 \\
			\bottomrule
		\end{tabular}
	}
\end{table}

The experiments included 1000 simulation runs and 20 real-world trials. The results are summarized in \tn{table5}. Experimental results show that our simulation model helps the algorithm to train effectively and achieve similar performance in the real environment. Our approach demonstrates significant advantages for cylindrical segment assembly with tight tolerance. Heuristic methods achieve high success rates in simulation due to noise-free conditions but drop significantly in real environments because of uncertainties. Spiral Search is limited to 300 steps. It has a large average step count and low efficiency. For LFD methods, BC achieves low success rates due to the limited expert dataset. DA-DMP is set with a time scaling factor $\tau=1$, and its average steps are consistent with the expert dataset. It yields satisfactory performance in simulation but degrades significantly in real‑world environments. Our method outperforms \cite{26} and \cite{27} in steps, reward and overall performance. All three methods exhibit good sim-to-real adaptability, while ours achieves the lowest performance degradation, showing the strongest robustness and generalization.

To further validate the generalization of our method to different assembly objects, experiments were conducted on a cylindrical segment with distinct geometric specifications. It features a $\Phi$200mm, a 0.15mm assembly tolerance, and 12 circumferential connecting holes, as shown in \fn{Fig_add2}. The policy adopted herein was directly transferred from the originally trained strategy, with the experimental setup consistent with the original trials.

\begin{figure}[h]
	\centering
	\includegraphics[width=0.4\columnwidth]{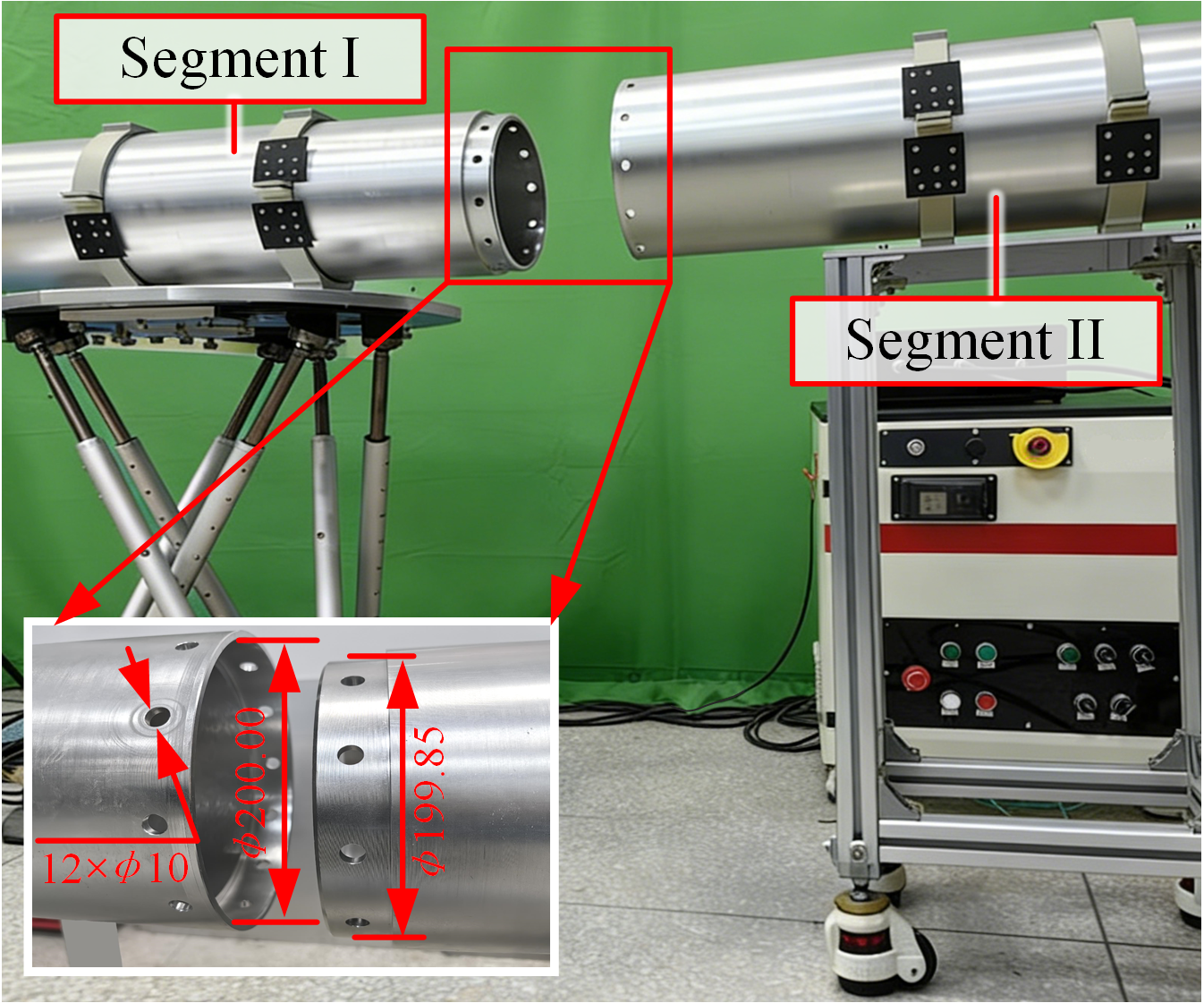}
	\caption{Experiment on the new assembly object.}
	\label{Fig_add2}
\end{figure}

\begin{table}[H]
	\footnotesize
	\centering
	\caption{Generalization performance comparison of different methods}
	\label{table6}
	\begin{tabular}{lccc} 
		\toprule
		Method & Avg. Steps & Avg. Reward & Success Rate \\
		\midrule
		\multicolumn{4}{l}{\textit{Heuristic Methods}} \\
		Visual Servoing \cite{22} & - & - & 0.80 \\
		Spiral Search \cite{add13} & 245 & - & 0.70 \\
		\midrule
		\multicolumn{4}{l}{\textit{Learning from Demonstration}} \\
		Behavior Cloning & 236 & 26.1 & 0.35 \\
		DA-DMP \cite{add15} & 190 & 49.8 & 0.55 \\
		\midrule
		Proposed method & 85 & 124.1 & 0.90 \\
		Method in \cite{27} & 117 & 87.9 & 0.80 \\
		Method in \cite{26} & 99 & 100.2 & 0.85 \\
		\bottomrule
	\end{tabular}
\end{table}

As shown in \tn{table6}, the proposed method maintains a high success rate of 0.9 after direct policy transfer with only slight performance degradation. Heuristic methods exhibit no significant performance changes, as all experimental conditions remain unchanged except for the assembly object, and their stability stems from reliance on fixed rule-based logic rather than data-driven adaptation. LFD methods show a noticeable performance decline due to the lack of new scenario data in the expert dataset. While the methods in \cite{26} and \cite{27} achieve basic assembly feasibility, their performance degradation is far more pronounced than that of the proposed method. The method in \cite{26} experiences a 11.1\% drop in success rate, and the method in \cite{27} a 15\% drop. This gap highlights the advantage of our framework design: the high-level LSTM dynamically adjusts action strategies based on real-time relative pose feedback, while the low-level BC-TD3 integrates and generalizes expert experience.

\section{Discussion}\label{Section: DISCUSSION}

The framework achieves high efficiency, high accuracy and better adaptability in both simulations and real-world scenarios. However, there are still limitations that need to be addressed in future work. Firstly, while our method demonstrates good adaptability to initial conditions beyond the training range and shows robustness to sudden disturbances in real-world environments, its robustness to physical parameter perturbations, changes in actuator precision, and noisy sensor conditions remains unexplored. These factors are critical in real-world industrial applications where environmental uncertainties and hardware imperfections are common. The binocular vision system’s accuracy directly affects assembly performance, but the correlation between its pose estimation error and assembly success lacks quantitative analysis. The system is untested under real-world visual disturbances, and simulated Gaussian noise fails to fully represent real non-Gaussian errors, potentially degrading practical performance. Secondly, the system relies solely on bounding box detection, which causes misjudgment and frequent collision-triggered emergency stops. Lacking force sensing and contact handling, unexpected collisions occur under tight tolerance, reducing safety and reliability in real industrial scenarios. Thirdly, the lack of specific energy efficiency metrics is a notable limitation. Although the smoother motion profiles of our method suggest potential energy savings, detailed measurements and evaluations of energy consumption during both learning and assembly phases are currently missing. Lastly, the current framework's scalability to more complex and larger-scale assembly tasks and its generalizability to a variety of assembly tasks have not been fully evaluated. While the framework demonstrates strong performance in the specific task of cylindrical segment assembly, its scalability to multi-component or more intricate assembly systems and its generalizability to different types of assembly tasks and industrial settings remain to be explored.

Future work will focus on addressing these limitations and expanding the framework's capabilities. We plan to conduct comprehensive evaluations of the framework's robustness by introducing physical parameter perturbations, varying actuator precision, and simulating noisy sensor conditions in real-world experiments. This will involve testing the framework under various practical scenarios to ensure its reliability and stability in the face of real-world uncertainties. We will also quantitatively analyze the relationship between vision estimation accuracy and assembly performance, and test system robustness against real-world visual disturbances. Moreover, a force-torque sensor will be integrated at the robot end-effector to enable real-time contact detection and active compliance control, thus reducing unexpected collisions and emergency stops. Additionally, we will incorporate detailed energy efficiency metrics into our evaluations. This will involve measuring and reporting specific energy consumption data during both learning and assembly phases, and exploring ways to optimize our method for energy efficiency. Furthermore, we aim to enhance both the scalability and generalizability of the framework. This will be achieved by extending its application to more complex and dynamic industrial scenarios, developing advanced control strategies to handle real-time changes and uncertainties, and applying it to multi-component and multi-hole alignment tasks. By exploring broader use cases, we will ensure the framework's broader applicability in real-world industrial settings, thereby enhancing its versatility and practical utility.

\section{Conclusion}\label{Section: CONCLUSION}
A hybrid hierarchical learning framework is proposed for the assembly of cylindrical segment components with tight tolerance. The task is divided into position and orientation adjustments, with dynamic step size adjustments breaking it into manageable subtasks. This reduces the complexity of high-dimensional spaces and improves learning efficiency. The lower-level network uses BC to integrate expert experience, giving the robot human-like intuition. Paired with the TD3 algorithm, the framework enhances training stability and robustness. The upper-level network dynamically adjusts lower-level decisions through LSTM based on BC-guided heuristic rules. Simulation environment is developed to learn before transferred to real world, for an efficient and safe learning process.


\section*{CRediT authorship contribution statement}
\textbf{Binbin Lian:} conducted the writing and editing of the manuscript. 
\textbf{Xinyu Liu:} carried out the methodology, software validation, data curation. 
\textbf{Tao Sun:} directed the project and conducted the writing and editing of the manuscript. 
\section*{Declaration of competing interest}
The authors declare that they have no known competing financial interests or personal relationships that could have appeared to influence the work reported in this paper.
\section*{Acknowledgments}
This work was supported in part by National Natural Science Foundation of China (NSFC) under grant No.52205028, in part by Tianjin Major Science and Technology Project on Artificial Intelligence under grant No. 25ZXRGGX00150, No. 25ZXRGGX00210 and No. 25ZXRGGX00200, and in part by the Beijing-Tianjin-Hebei Fundamental Research Project under grant No. E2024203249, and in part by CAST Innovation Fund Project under grant No. QF-2025-00074, and in part by Xiaomi Young Talent Program, and in part by Guizhou Provincial Science and Technology Support Project under grant No. 2025-136 and in part by Tianjin Science and Technology Plan Project under grant No. 24JCZXJC00240, and in part by Seed Foundation of Tianjin University under grant No. 2025XJ1-0012. 
\section*{Data availability}
All data are provided in the manuscript and the Supplementary Materials.

\bibliographystyle{elsarticle-num}
\bibliography{bibliography.bib}
\end{document}